%% file: main.tex
\documentclass[]{jingdong}
\input{math_commands.tex}

\usepackage{float}
\usepackage{url}
\usepackage{amssymb}
\usepackage{tabularx}
\usepackage{array}
\usepackage{xspace}
\usepackage{listings}
\usepackage{enumitem}
\usepackage{etoc}

\microtypesetup{expansion=false}
\renewcommand{\bibfont}{\small}
\newcommand{\camg}{\textsc{CAMG}\xspace}
\newcommand{\camgrl}{\textsc{CAMG-RL}\xspace}
\newcolumntype{L}[1]{>{\raggedright\arraybackslash}p{#1}}

\definecolor{casepromptblue}{RGB}{31,86,140}
\definecolor{casemetagray}{RGB}{92,100,110}
\definecolor{caseboxbg}{RGB}{247,248,250}
\definecolor{caseboxrule}{RGB}{200,205,212}
\definecolor{casetitlebar}{RGB}{31,86,140}
\newcommand{\casetitle}[2]{%
  \par\noindent\colorbox{casetitlebar}{\parbox{\dimexpr\linewidth-2\fboxsep}{%
    \strut\color{white}\ttfamily\small\textbf{#1}\hfill\normalfont\itshape\footnotesize #2\strut}}%
  \nopagebreak\par\nointerlineskip\vspace{1pt}}
\lstdefinestyle{taskExcerpt}{
  basicstyle=\ttfamily\scriptsize,
  breaklines=true,
  breakatwhitespace=false,
  columns=fullflexible,
  keepspaces=true,
  showstringspaces=false,
  backgroundcolor=\color{caseboxbg},
  frame=single,
  rulecolor=\color{caseboxrule},
  framexleftmargin=3pt,
  xleftmargin=3pt,
  xrightmargin=3pt,
  aboveskip=0pt,
  moredelim=[l][\color{casepromptblue}\bfseries]{[TASK INPUT]},
  moredelim=[l][\color{casemetagray}\bfseries]{[NATIVE INTERACTION]},
  moredelim=[l][\color{casemetagray}\bfseries]{[SHARED FILE INTERFACE]},
  moredelim=[l][\color{casemetagray}\bfseries]{[VERIFICATION]},
  moredelim=[l][\color{casepromptblue}\bfseries]{[FILE-MEMORY WRITE]},
  moredelim=[l][\color{casepromptblue}\bfseries]{[FILE-MEMORY READ]},
  moredelim=[l][\color{casemetagray}\bfseries]{[CONTEXT REPLACEMENT]},
  moredelim=[l][\color{casemetagray}\bfseries]{[NATIVE OUTCOME]}
}
\newcounter{algorithm}

\title{Coding Agent Memory Post-training:\\Unlocking the Memory Potential of\\Pre-trained File Operations\\for Long-Horizon Tasks\\via Reinforcement Learning}
\author{Lirui Luo}
\author{Kelong Mao}
\author{Heming Xia}
\author{Rongqing Li}
\author{Xinwei Yang}
\author{\protect\\[+0.25em]Luyu Chen}
\author{Kieran Wong}
\author{Yudong Guo}
\author{Xinrui Wang}
\author{\protect\\[+0.25em]Jiayin Zhu}
\author{Simiu Gu}
\author{Sulong Xu}
\author{Cong Fang}
\affiliation{JD.com}
\checkdata[Project page]{\url{https://liruiluo.github.io/agentmemorygym/}}

\abstract{\input{sections/abstract.tex}}

\begin{document}
\raggedbottom
\maketitle

\input{sections/intro.tex}
\FloatBarrier
\input{sections/related.tex}

\FloatBarrier
\input{sections/preliminaries.tex}
\input{sections/camg.tex}
\input{sections/camg_rl.tex}
\FloatBarrier
\input{sections/experimental_setup.tex}
\input{sections/results.tex}

\FloatBarrier
\input{sections/discussion.tex}
\FloatBarrier

\input{sections/statements.tex}

\bibliography{references}
\bibliographystyle{iclr2027_conference}

\clearpage
\appendix
\makeatletter
\providecommand{\protected@file@percent}{}
\def\addcontentsline#1#2#3{%
  \begingroup
    \let\label\@gobble
    \@ifundefined{@currentHref}{\def\@currentHref{}}{}%
    \addtocontents{#1}{\protect\contentsline{#2}{#3}{\thepage}{\@currentHref}\protected@file@percent}%
  \endgroup}
\makeatother
\input{sections/appendix.tex}

\end{document}

%% file: math_commands.tex
\usepackage{amsmath,amsfonts,bm}

\def\eqref#1{equation~\ref{#1}}

\def\1{\bm{1}}

\DeclareMathAlphabet{\mathsfit}{\encodingdefault}{\sfdefault}{m}{sl}
\SetMathAlphabet{\mathsfit}{bold}{\encodingdefault}{\sfdefault}{bx}{n}



%% file: sections/abstract.tex
Language-model agents increasingly tackle long-horizon tasks whose interaction histories exceed the
model's active context. Recent work has begun to use reinforcement learning to make memory control
part of the policy, often relying on predefined memory tools within domain-specific training
environments of relatively short horizons. This setup ties learned memory behavior to
environment-specific interfaces that lie outside the base model's pre-training and must be learned
from scratch, so even after post-training, agents struggle to use memory in long-horizon tasks. To
address these limitations, we introduce \textbf{Coding Agent Memory Gym (\camg)}, a suite of
long-horizon agentic-RL environments spanning Shop, Coding, DeepResearch, and AutoResearch.
Alongside each environment's native task interface, CAMG provides executable shell access and an
episode-persistent workspace, enabling agents to create, revise, search, and reuse files as memory
throughout an episode. We also introduce \textbf{\camgrl}, which trains a single policy jointly
across all four environments with fully asynchronous PPO, learning this file-based memory behavior
directly from downstream task reward, and we train \mbox{\camgrl-4B} and \mbox{\camgrl-9B} from
Qwen3.5 models of matching size. On SWE-bench Verified and MLE-bench Lite, \mbox{\camgrl-4B} and
\mbox{\camgrl-9B} are competitive with Qwen3.5-35B-A3B and Qwen3.5-122B-A10B, respectively.

%% file: sections/intro.tex
\section{Introduction}
\label{sec:intro}

Language-model agents increasingly tackle long-horizon tasks that unfold through repeated tool use and
environment interaction \citep{yang2024sweagent,agentgymrl2025,gao2026unlocking}. As these interactions
accumulate, information needed for later decisions can fall out of the model's active context. A coding
agent, for example, may need to recall why an earlier fix was rejected after that exchange has left the
prompt.

Recent work has begun to use reinforcement learning to make memory control part of the policy, often
relying on predefined memory tools within domain-specific training environments of relatively short
horizons \citep{yan2026memoryr1,huo2026atommem,liu2026pensieve,yu2026agenticmemory,li2026compactionrl}.
This setup ties learned memory behavior to environment-specific interfaces that lie outside the base
model's pre-training, so even after post-training, agents struggle to use memory in long-horizon tasks. What
remains open is whether reinforcement learning on top of code pre-training can elicit a coding agent's
ability to retain and reuse information beyond its active context.

To address these limitations, we introduce \textbf{Coding Agent Memory Gym (\camg)}, a suite of
long-horizon agentic-RL environments spanning Shop, Coding, DeepResearch, and AutoResearch. Across all
four environments, CAMG pairs each native task interface with executable shell access and an
episode-persistent workspace. Agents can therefore create, revise, search, and reuse files as memory
throughout an episode.

\begin{figure}[H]
  \centering
  \includegraphics[width=0.96\linewidth]{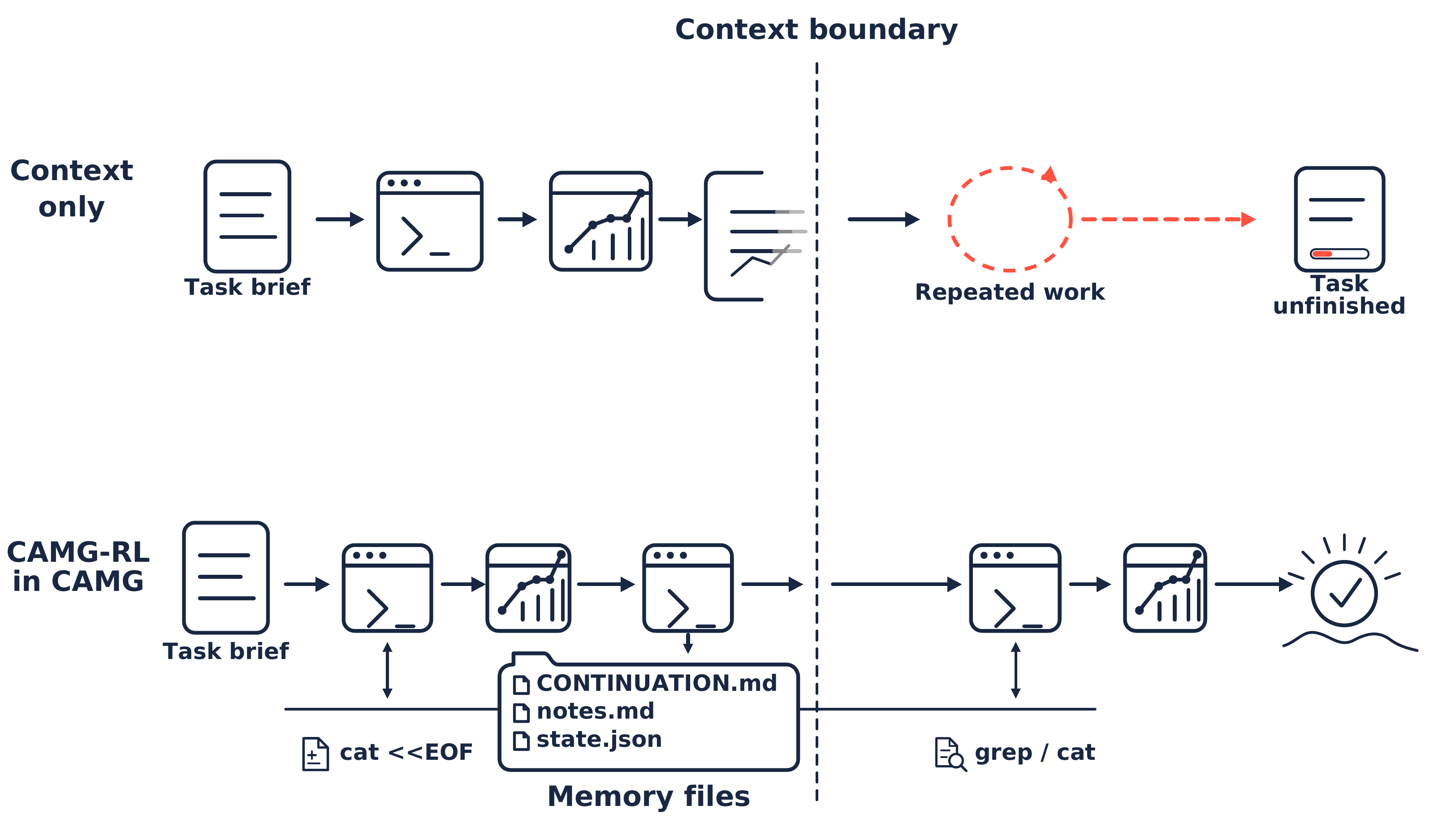}
  \caption{\textbf{CAMG-RL learns to use files as memory for long-horizon tasks.}
  Each CAMG environment provides the task-acting policy with an episode-persistent workspace that it can
  write, revise, search, and read through ordinary filesystem actions throughout the episode. The illustrated
  context-boundary case shows the policy updating a bounded \texttt{CONTINUATION.md} before context
  replacement and retrieving the saved working state through an ordinary file read afterward.}
  \label{fig:camg-teaser}
\end{figure}

We also introduce \textbf{\camgrl}, which jointly trains a single task-acting policy across all four
environments, learning from downstream task reward through this common interface, as
Figure~\ref{fig:camg-teaser} illustrates. The policy emits executable actions, may open each response
with free-form reasoning, and develops its long-horizon memory state in persistent, editable files. These
files retain plans, evidence,
and intermediate findings across context replacements, while task reward shapes their creation and
reuse. We implement CAMG-RL with fully asynchronous PPO so rollout and learning proceed concurrently
across variable-duration trajectories, and we train \mbox{\camgrl-4B} and \mbox{\camgrl-9B} from Qwen3.5 models of matching size.

Our contributions are threefold:
\begin{enumerate}
  \item \textbf{A long-horizon agentic-RL environment suite.} We introduce CAMG with four environments
  that require memory across context boundaries, each pairing native task interfaces with executable
  shell access and an episode-persistent workspace.
  \item \textbf{A reinforcement-learning algorithm for coding-agent memory.} We introduce \mbox{CAMG-RL},
  which learns from downstream task reward through fully asynchronous PPO to use ordinary coding-agent
  filesystem operations as memory.
  \item \textbf{A systematic empirical study of learned coding-agent memory.} We show that
  reinforcement learning on top of code pre-training can elicit reusable long-horizon memory:
  \mbox{\camgrl-4B} achieves the highest average success rate across the four test environments and, on
  SWE-bench Verified and MLE-bench Lite, \mbox{\camgrl-4B} and \mbox{\camgrl-9B} are competitive with
  Qwen3.5-35B-A3B and Qwen3.5-122B-A10B, respectively.
\end{enumerate}

%% file: sections/related.tex
\section{Related Work}
\label{sec:related}

\paragraph{Learned memory control.}
These methods let the policy decide when and how to store, update, and retrieve information.
Earlier work trains the policy to use dedicated operations over external stores
\citep{yan2026memoryr1,huo2026atommem}, and later work extends this control to shorter-lived working or
episodic state \citep{liu2026pensieve,yu2026agenticmemory,sun2026vermem}. CAMG-RL introduces no special
memory tools: on top of code pre-training, it elicits the task-acting coding agent's ability to maintain
memory through ordinary filesystem actions.

\paragraph{Learning to compact context.}
These methods optimize the information retained across a bounded-context transition.
\citet{kang2026acon} and \citet{li2026contextcurator} train a separate manager to compress context for a
fixed task executor. Other architectures place bounded internal state inside the
trained policy \citep{li2026mempo,yuan2026memsearcher,zhou2026mem1,yu2026memagent}.
Policy-generated summaries or adaptive compression also receive direct supervision from downstream outcomes
\citep{lu2026beyond,li2026compactionrl,hu2026ziprl}. Across these architectures, the retained state lives
in the active context or inside the trained policy. CAMG-RL instead retains it in a bounded continuation
file at each context replacement.

\paragraph{Long-horizon agentic-RL training environments.}
These environments train agents through executable feedback over extended interactions.
One environment suite spans diverse interactive tasks through a common interface \citep{xi2024agentgym},
and subsequent work adds multi-turn policy optimization and longer interaction horizons
\citep{agentgymrl2025}. Other environments
score repository repairs with runnable tests \citep{pan2025swegym}, evaluate iterative improvement of
machine-learning artifacts \citep{nathani2025mlgym,mledojo2025}, or benchmark budget-constrained basket
shopping \citep{li2026comboshoppingbench}. CAMG provides diverse, verifiable
environments across four task families that require memory across context boundaries, each pairing its
native task interface with executable shell access and an episode-persistent workspace.

Table~\ref{tab:positioning} summarizes these distinctions. Appendix~\ref{app:extended-related-work}
discusses additional related work.

\input{tables/positioning_comparison_table}

%% file: tables/positioning_comparison_table.tex
\begin{table}[H]
  \centering
  \footnotesize
  \caption{\textbf{CAMG uniquely enables end-to-end RL of filesystem memory across verifiable, cross-context tasks in multiple domains.}
  Each environment produces its task reward by a fixed procedure from executed policy behavior: three use
  programmatic checks, and DeepResearch answers are scored by a frozen semantic judge shared by every
  evaluated method (Appendix~\ref{app:literesearcher}).
  We position CAMG among agent-memory benchmarks and trainable agent environments.
  \checkmark{}~= present, $\times$~= absent.}
  \label{tab:positioning}
  \setlength{\tabcolsep}{1.2pt}
  \renewcommand{\arraystretch}{1.06}
  \begin{tabularx}{\linewidth}{L{0.38\textwidth} *{5}{>{\centering\arraybackslash}X}}
    \toprule
    & {\scriptsize\shortstack{Multiple\\domains}}
    & {\scriptsize\shortstack{Task\\reward}}
    & {\scriptsize\shortstack{End-to-end\\RL}}
    & {\scriptsize\shortstack{Task across\\contexts}}
    & {\scriptsize\shortstack{Shell\\access}} \\
    \midrule
    \multicolumn{6}{l}{\scriptsize\textit{Agent-memory benchmarks}} \\
    LongMemEval \citep{wu2025longmemeval}
      & $\times$ & $\times$ & $\times$ & $\times$ & $\times$ \\
    MemoryAgentBench \citep{hu2026memoryagentbench}
      & $\times$ & $\times$ & $\times$ & $\times$ & $\times$ \\
    MemoryArena \citep{he2026benchmarking}
      & \checkmark & $\times$ & $\times$ & \checkmark & $\times$ \\
    \addlinespace[1pt]
    \multicolumn{6}{l}{\scriptsize\textit{Trainable long-horizon agent environments}} \\
    AgentGym-RL \citep{agentgymrl2025}
      & \checkmark & \checkmark & \checkmark & $\times$ & $\times$ \\
    SWE-Gym \citep{pan2025swegym}
      & $\times$ & \checkmark & $\times$ & $\times$ & \checkmark \\
    MLE-Dojo \citep{mledojo2025}
      & \checkmark & \checkmark & $\times$ & $\times$ & $\times$ \\
    \midrule
    \textbf{Coding Agent Memory Gym (ours)}
      & \checkmark & \checkmark & \checkmark & \checkmark & \checkmark \\
    \bottomrule
  \end{tabularx}
\end{table}

%% file: sections/preliminaries.tex
\section{Preliminaries}
\label{sec:preliminaries}

\noindent\textbf{Multi-turn agentic reinforcement learning.}\spacefactor=1000\space\space%
We study multi-turn interactive decision-making tasks,
i.e., agentic tasks, and model each environment $d\in\mathcal{D}$ as a finite-horizon partially
observable Markov decision process \citep{sutton2018reinforcement},
$\mathcal{M}_d=(\mathcal{U}_d,\mathcal{S}_d,\mathcal{A}_d,\mathcal{O}_d,\mathcal{T}_d,r_d)$.
These terms denote the instruction, latent-state, action, and observation spaces, together with the
state-transition and reward functions. Given an instruction $u$ drawn from the environment's
instruction distribution $\mu_d$, at turn $t$ the policy observes the visible interaction history
$h_t$ and generates one complete response
$a_t\sim\pi_\theta(\cdot\mid h_t)$. The environment executes that response as an action, returns an
observation, and transitions to the next state. The actions and observations collected over at most
$H_d$ turns form a trajectory $\tau$ of realized length $T(\tau)$ with task return
$G_d(\tau)=\sum_{t=1}^{T(\tau)}\gamma^{t-1}r_{d,t}$, where $\gamma$ is the discount factor.

\noindent\textbf{Policy gradient.}\spacefactor=1000\space\space%
The policy maximizes expected task return,
\begin{equation}
  J_d(\theta)=\mathbb{E}_{\tau\sim p_{\theta,d}}[G_d(\tau)],
  \label{eq:policy-gradient-objective}
\end{equation}
where $p_{\theta,d}$ is the trajectory distribution induced by the policy and environment. The
vanilla policy gradient is
\begin{equation}
  \nabla_\theta J_d(\theta)
  =\mathbb{E}_{\tau\sim p_{\theta,d}}
  \left[G_d(\tau)\sum_{t=1}^{T(\tau)}
  \nabla_\theta\log\pi_\theta(a_t\mid h_t)\right].
  \label{eq:policy-gradient}
\end{equation}

%% file: sections/camg.tex
\section{Coding Agent Memory Gym}
\label{sec:camg}

To train agents to use files as memory, we construct CAMG as a suite of four long-horizon environments.
Each environment combines verifiable tasks with executable shell access and an episode-persistent
workspace. Their tasks require agents to reuse information accumulated earlier in the episode.
Appendix~\ref{app:training-envs} describes the construction, selection, and data splits of all four
environments.

\subsection{Constructing Four Long-Horizon Environments}
\label{sec:environment-construction}

\paragraph{Shop.}
Shop tests whether an agent can retain customer requirements and earlier purchase decisions across a
sequence of dependent purchases. We generate six-session tasks from 720 verified WebShop
products \citep{yao2022webshop}. Between sessions the environment clears the active
session transcript, so later purchases depend on what the policy retains.

\paragraph{Coding.}
Coding tests whether an agent can complete a software repair whose diagnosis, repository edits, and test
evidence may span multiple active contexts. We draw issue--repository pairs and test-defined repair objectives from SWE-smith
\citep{yang2025swesmith}. We exclude non-executable tasks and those whose reference repairs fail
their tests, retaining 11,218 tasks from 93 repositories.

\paragraph{DeepResearch.}
DeepResearch tests whether an agent can accumulate and reuse retrieved evidence over an extended search.
We use LiteResearcher's Search--Visit--Answer interaction to gather evidence from a fixed retrieval
corpus before a scored final answer
\citep{literesearcher2025}. We exclude tasks that cannot run or be judged, combining the two
LiteResearcher releases into 26,597 tasks.

\paragraph{AutoResearch.}
AutoResearch tests whether an agent can preserve hypotheses and validation evidence across iterative model
development. We use public OpenMLE task packages with datasets, executable submissions, and native metrics
\citep{yang2026frontisma1}. We exclude tasks that cannot run or whose reference pipelines fail,
retaining 931 tasks from 804 task families.

\subsection{A Common Coding-Agent Interface}
\label{sec:interface-and-families}

Across the four environments, CAMG standardizes an executable filesystem interface for using files as
memory while retaining each environment's native task interaction and reward. Following the minimal,
shell-only interface of mini-swe-agent \citep{yang2024sweagent}, the shared filesystem action is a single
\texttt{shell\_command}, which can create, inspect, search, modify, and execute files, so the policy can
preserve and revise task-relevant evidence and intermediate state in the workspace. Let
$\mathcal A_d^{\mathrm{task}}$ denote environment $d$'s native task-action set and
$\mathcal A^{\mathrm{fs}}$ the shared filesystem-action set. The CAMG action space is
\begin{equation}
  \mathcal A_d=\mathcal A_d^{\mathrm{task}}\cup\mathcal A^{\mathrm{fs}},
  \qquad a_t\in\mathcal A_d.
  \label{eq:camg-action-space}
\end{equation}
One \texttt{shell\_command} can produce both native-task and filesystem effects, so the two sets may
overlap. Appendix~\ref{app:state-action-contract} specifies the common interface.

\begin{figure}[!ht]
  \centering
  \includegraphics[width=\linewidth]{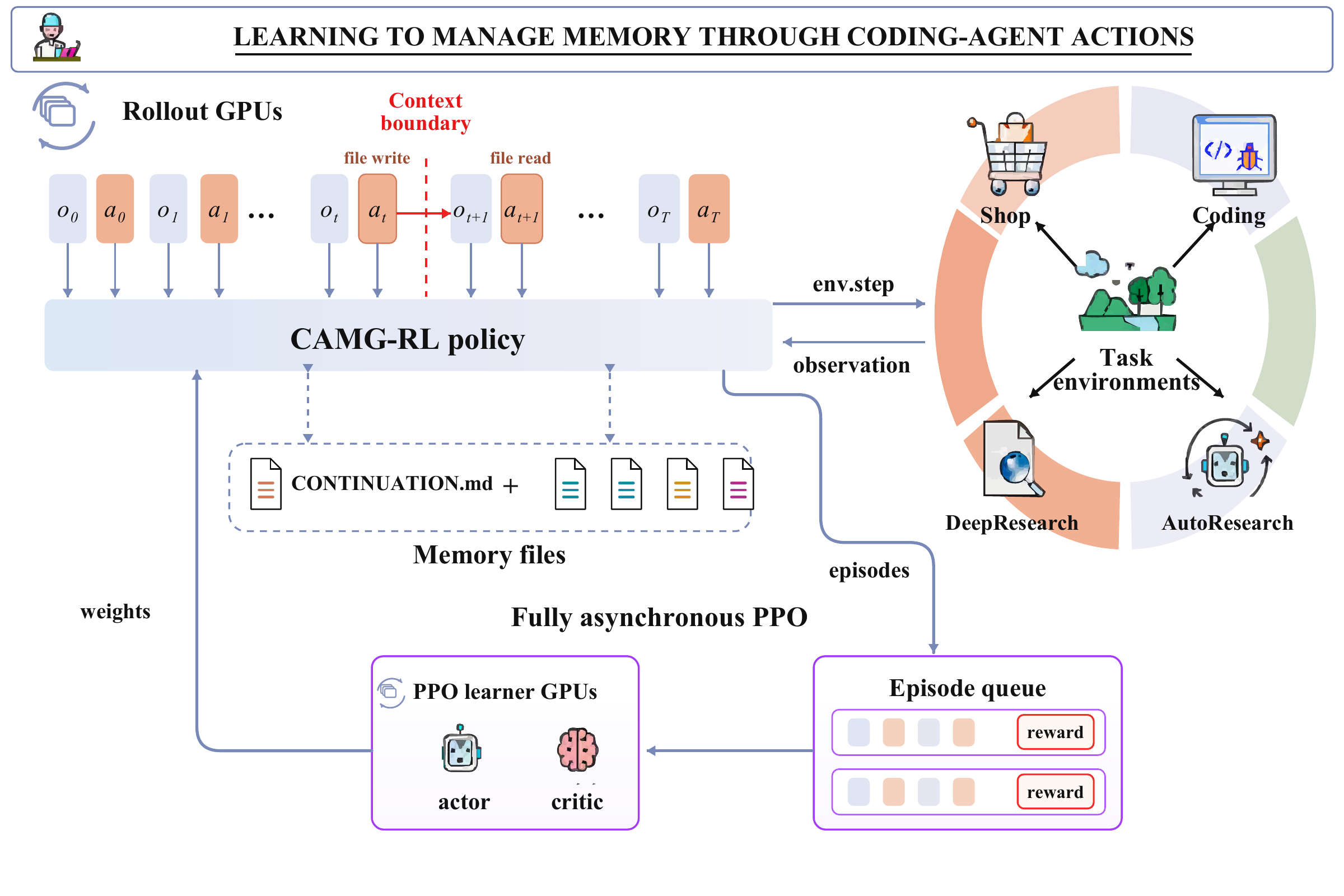}
  \caption{\textbf{A single policy learns to use files as memory while solving long-horizon tasks.}
  Task interactions and filesystem operations form one policy trajectory over an episode-persistent
  workspace. At the highlighted context boundary, the policy updates \texttt{CONTINUATION.md}.
  A later read brings the saved working state back into active context. Each completed episode then
  carries its full trajectory record into fully asynchronous PPO, where the learner consumes queued
  episodes and publishes updated weights for later episodes.}
  \label{fig:camg-framework}
\end{figure}

Figure~\ref{fig:camg-framework} summarizes how this common interface connects task interaction and file
use to the learning pipeline described next.

%% file: sections/camg_rl.tex
\section{CAMG-RL}
\label{sec:camg-rl}

Section~\ref{sec:camg} equips every task with executable filesystem actions over an episode-persistent
workspace. We now ask how a policy can learn to use this workspace as memory from downstream task
outcomes.

\subsection{Learning to Use Files as Memory}
\label{sec:camg-rl-learning}

At each turn, the policy samples $a_t\in\mathcal A_d$ as defined in Eq.~\ref{eq:camg-action-space}.
Each response may open with a free-form reasoning span and always ends with an executable action. We
place task interactions and filesystem
operations on the same action trajectory, propagate downstream rewards with GAE \citep{schulman2016gae}, and optimize every
sampled token, including policy-generated file contents, with clipped PPO under the task-return
objective in Eq.~\ref{eq:policy-gradient-objective} \citep{schulman2017ppo}.

\paragraph{Dual-granularity PPO.} Existing multi-turn RL
systems assign both operations a single granularity, turn-level in Turn-PPO \citep{li2025turnppo}
and token-level in SAO \citep{hou2026sao}. Turn-level estimation is stable, because one reward
arrives per response, but coarse, because every token in a response shares one advantage.
Token-level estimation is fine-grained but high-variance, since splitting that reward across tokens
depends on a learned value function. CAMG-RL therefore estimates one advantage per response and
clips one ratio per sampled token, keeping the estimation stable and the update fine-grained.

This learning setup
treats persistent, editable files as the policy's long-horizon memory state. For the four-environment set
$\mathcal D$, each rollout samples an
environment uniformly, giving the joint objective
\begin{equation}
  J_{\textsc{CAMG-RL}}(\theta)
  =\frac{1}{|\mathcal D|}\sum_{d\in\mathcal D}J_d(\theta),
  \qquad |\mathcal D|=4.
  \label{eq:uniform-environment-mixture}
\end{equation}
Appendix~\ref{app:camg-rl-procedure} defines the policy trajectory. Credit-assignment details
appear in Appendix~\ref{app:action-credit}.

\subsection{Fully Asynchronous PPO for Variable-Duration Episodes}
\label{sec:fully-async-implementation}

Episode duration varies across CAMG because rollouts take different numbers of turns and environment
tools have different execution costs. In a synchronous rollout batch, every worker waits for the slowest
episode. We therefore run rollout and learning fully asynchronously. One rollout process collects
trajectories from all four environments. Rollout workers append completed episodes to a persistent
queue, while the learner assembles them into optimization batches, updates one shared actor and critic, and
makes the updated actor available to later episodes.
The queue therefore absorbs uneven episode completion times, so no worker waits for the slowest
episode.
Appendix~\ref{app:fully-async-algorithm} gives the execution algorithm, and Appendix~\ref{app:training-details}
gives the training configuration.

%% file: sections/experimental_setup.tex
\section{Experiments}
\label{sec:experimental-setup}

We first compare CAMG-RL with baselines on the test set and explain why filesystem memory
makes effective use of pre-training. We then measure external transfer and ablate the
component variants. Finally, we isolate the causal role of saved memory and present a case
study.

\subsection{Experimental Setup}
\label{sec:setup}

\noindent\textbf{Training.}\spacefactor=1000\space\space%
All trainable methods use Qwen3.5-4B \citep{qwen2026qwen35} for 200 learner optimizer updates with 64 episodes per update, totaling 12,800 episodes, under the uniform mixture over Shop, Coding, DeepResearch, and AutoResearch in Eq.~\ref{eq:uniform-environment-mixture}, trained with fully asynchronous PPO and native task rewards.
Appendix~\ref{app:learning-systems} gives the remaining hyperparameters and runtime details.

\noindent\textbf{Baselines.}\spacefactor=1000\space\space%
As training-free controls on the frozen Qwen3.5-4B base, we evaluate Mem0 \citep{chhikara2025mem0}, which manages an external store at inference, and Letta Code \citep{letta2026lettacode}, whose git-backed MemFS descends from MemGPT \citep{packer2023memgpt}. Trainable baselines differ only in memory, keeping learned context compaction for CompactionRL \citep{li2026compactionrl} and six LTM/STM tools for AgeMem \citep{yu2026agenticmemory}.

\noindent\textbf{Evaluation.}\spacefactor=1000\space\space%
We report per-environment success rates and their equal-weight average, with the success definitions in Appendix~\ref{app:evaluation-details}.
Trainable methods are trained using the CAMG train split and evaluated using the CAMG test set, and all methods use the same 128 test tasks per environment, giving 512 tasks in total, with matched greedy decoding, action budgets, runtimes, and graders.
Unless stated otherwise, all frozen models use CAMG-RL's memory mechanism.
We measure external transfer on SWE-bench Verified and MLE-bench Lite~\citep{jimenez2024swebench,chan2024mlebench}, where all rows use the same tasks, decoding, action budgets, task environments, and graders, and unsuccessful assigned tasks count as failures.

%% file: sections/results.tex
\subsection{Results}
\label{sec:results}

\input{tables/native_heldout_result_table}

\paragraph{Learning filesystem memory beats training-free memory systems, learned compaction, and tool-based memory.}
CAMG-RL achieves the highest average success rate, 54.4\% versus 48.0\% for CompactionRL, the
strongest baseline, because its memory files avoid the information loss that learned compaction
incurs (Table~\ref{tab:native-heldout}). On Shop both methods nearly saturate the environment (97.1\% versus
97.5\%), and the average gap opens on Coding and AutoResearch.
AgeMem trails far behind, perhaps because its dedicated tools fall outside the pre-training
distribution and are hard for the model to use.
Training-free memory systems fare no better: the frozen 4B policy they augment has
a weak prior over when to store, retrieve, and use memory. Added storage leaves the task bottleneck in place (Appendix~\ref{app:action-accounting}
reports the full action accounting).

\begin{figure}[!ht]
  \centering
  \includegraphics[width=0.98\linewidth]{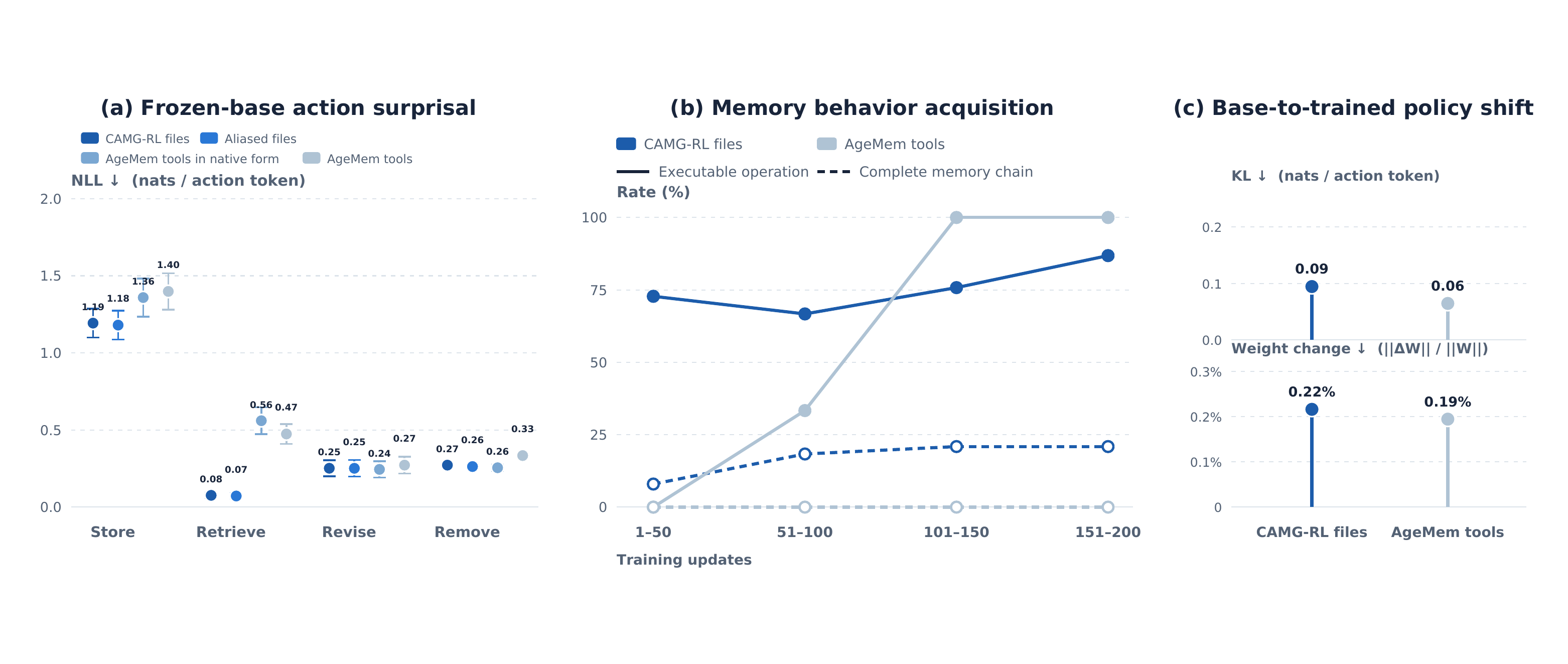}
  \caption{\textbf{The frozen base already prefers filesystem memory actions, and training converts
  this preference into complete memory chains while barely moving the weights.} (a)~Frozen-base action
  surprisal on paired store, retrieve, revise, and remove intents. Renaming the filesystem tool
  to an unseen name (\emph{Aliased files}) keeps the surprisal within 0.013 nats per token of the
  native actions, and the hybrid control (\emph{AgeMem tools in native form}) recovers only a fifth
  of the store gap.
  (b)~Complete store--retrieve--use chains and executable operations during training. AgeMem never
  completes a chain. (c)~Mean token-averaged KL of each trained policy from the frozen base, and
  relative weight change over the language-model weights.}
  \label{fig:interface-prior}
\end{figure}

\paragraph{The base policy already prefers filesystem memory actions over dedicated memory tools.}
To test the hypothesis that the dedicated tools fall outside the pre-training distribution, we
measure how surprising each memory action is to the frozen base, the mean negative log
probability of the action tokens in nats per token
(Figure~\ref{fig:interface-prior}). Filesystem memory actions cost 0.076
to 1.193 nats per token across the paired store, retrieve, revise, and remove intents, while
AgeMem's six dedicated tools are more surprising on every intent in their own call format. In
Table~\ref{tab:interface-name-cost} we further rule out the tool names, the call format, and the
documented shell command as explanations for the gap. The prior
therefore favors the filesystem action itself.

\paragraph{RL training on filesystem memory unlocks pre-trained capabilities that dedicated memory tools leave untapped.}
Does RL training build on this prior? We track complete store--retrieve--use chains
over the course of training (Figure~\ref{fig:interface-prior}b). These chains count only memory use
beyond the continuation write that every context replacement already requires, so the reported chain
rates are voluntary use on top of that mandatory write. CAMG-RL's chains rise from 8.0\% of
episodes in the first quarter to 20.9\% in the last, while AgeMem attempts its memory tools in
53 of 261{,}941 responses and never completes a chain. To check whether this difference reflects how
far training moved each policy, we measure the distance of each trained policy from the frozen
base, in action distribution and in weight space (Figure~\ref{fig:interface-prior}c). Both distances
are small, yet only CAMG-RL converts this movement into complete memory chains:
training on filesystem memory makes effective use of pre-training. Appendix~\ref{app:interface-prior-analysis} gives the complete protocol.

\input{tables/matched_external_method_comparison_table}

\paragraph{External benchmarks show trained task competence under distribution shift.}
Whether the trained competence transfers beyond the CAMG training distribution remains open. We evaluate the final
CAMG-RL-4B and CAMG-RL-9B checkpoints on SWE-bench Verified and MLE-bench Lite, which cover the Coding
and AutoResearch domains. CAMG-RL-4B is competitive with Qwen3.5-35B-A3B and CAMG-RL-9B with the
frozen Qwen3.5-122B-A10B on both (Table~\ref{tab:matched-external-methods}). On SWE-bench Verified,
CAMG-RL-4B resolves 15.8\% of assigned tasks, between the frozen 9B and 27B, and CAMG-RL-9B resolves
27.6\%, above the frozen 122B-A10B. On MLE-bench Lite they reach 4.5\% and 9.1\%, CAMG-RL-4B matching
the frozen 9B, 27B, and 35B-A3B and CAMG-RL-9B matching the frozen 122B-A10B.

\begin{figure}[H]
  \centering
  \includegraphics[width=0.98\linewidth]{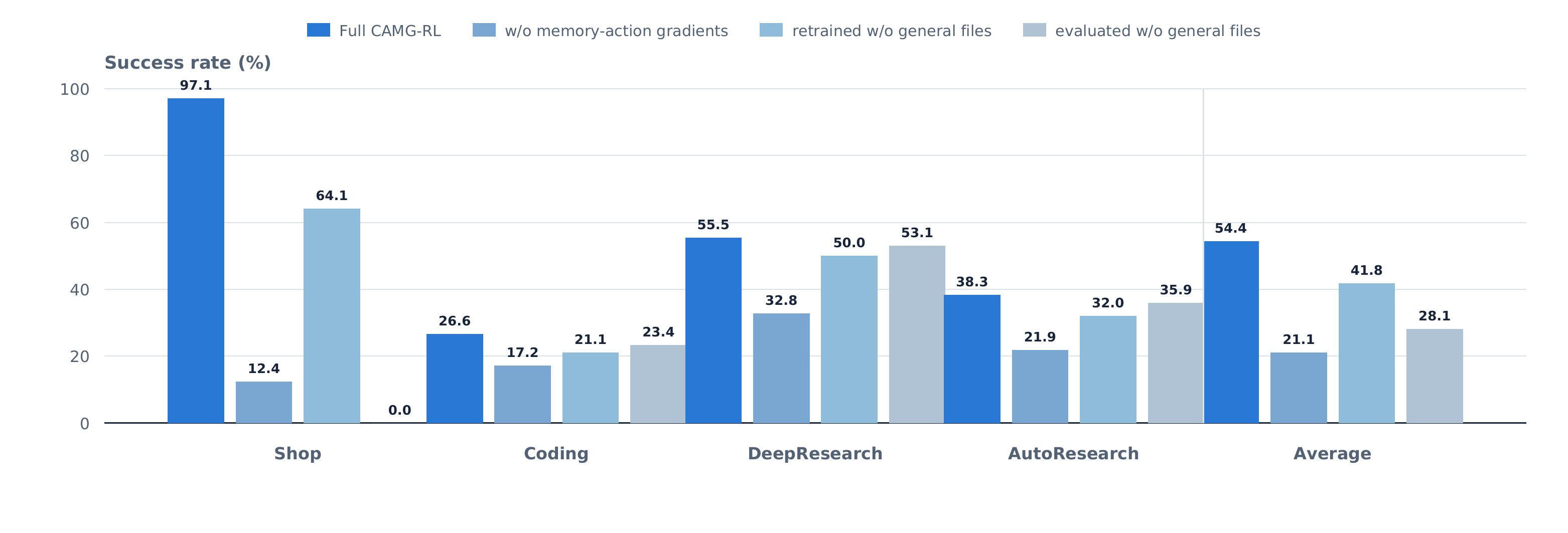}
  \caption{\textbf{Memory-action gradients and general memory files both drive CAMG-RL's gains, and the
  evaluation-only restriction trades Shop for the other three environments.} Grouped bars report the four environment success
  rates and their equal-weight average for full CAMG-RL, w/o memory-action gradients, retrained
  w/o general files, and evaluated w/o general files on the
  same test tasks.}
  \label{fig:camg-rl-ablation-expected}
\end{figure}

\paragraph{Memory-action gradients and general memory files both drive CAMG-RL's gains.}
We attribute the gains by isolating two parts of the method: direct actor learning from memory-file responses and access to
general memory files beyond the bounded continuation file. Figure~\ref{fig:camg-rl-ablation-expected}
compares full CAMG-RL with three component variants, one for the gradients and two for the memory files. The variant without memory-action gradients removes
memory responses from the actor loss and nothing else. The variant retrained without general memory files is a separately initialized policy trained with the
continuation file as its only memory file. The variant evaluated without general memory files restricts the
fully trained policy to that file without further training. Removing memory-action gradients
drops the average to 21.1\%, close to the frozen base. Training and evaluating with only the
continuation file reaches an average of 41.8\% versus 54.4\% for the full policy.

\paragraph{Saved memory content drives later success.}
The saved content could accompany success without driving it. We therefore intervene on a trajectory's saved
memory files at its first context replacement. Blanking them cuts average success by 14.1 points, and
transplanting task-mismatched content cuts it by 11.5, with the largest drop in Shop
(Appendix~\ref{app:memory-content-intervention}, Figure~\ref{fig:memory-intervention}).

\begin{figure}[H]
  \centering
  \includegraphics[width=0.98\linewidth]{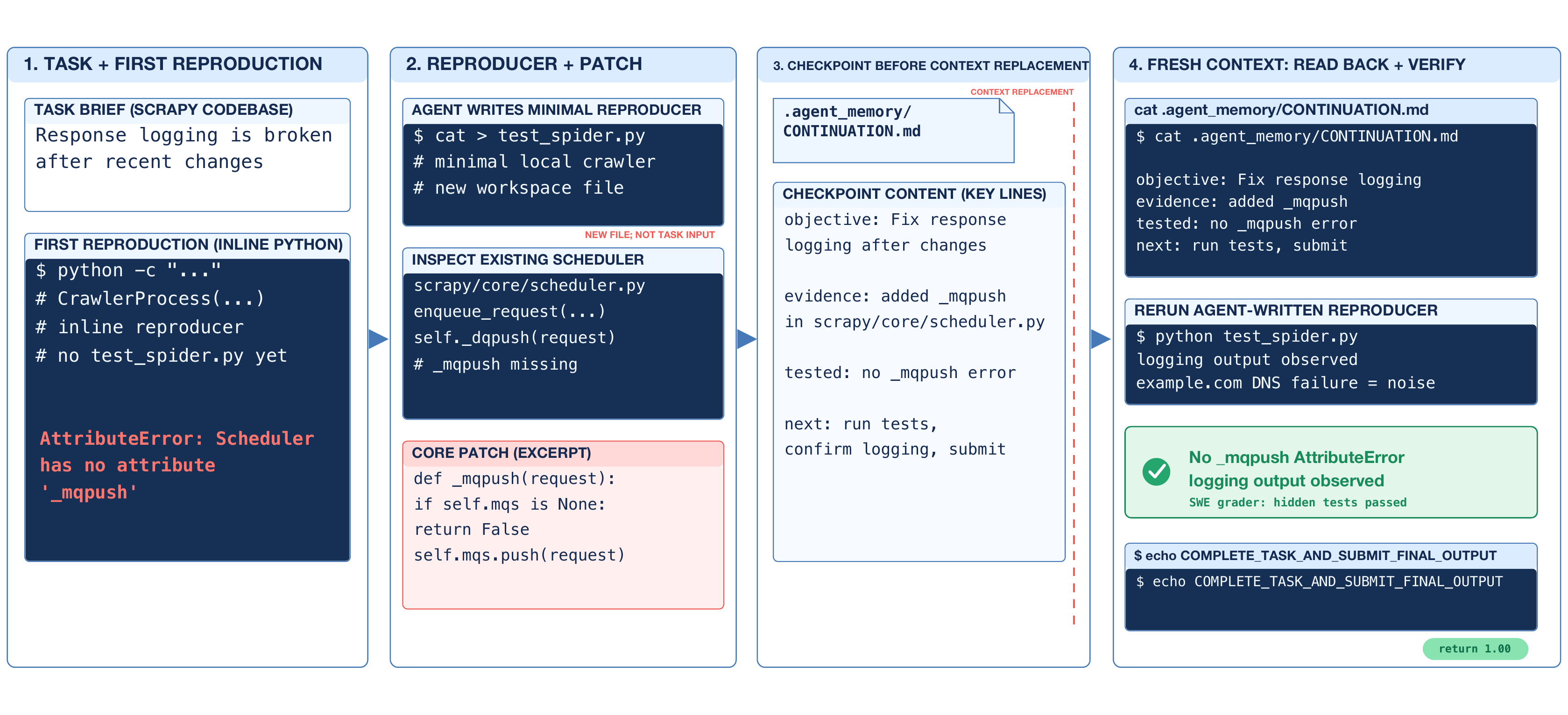}
  \caption{\textbf{The trained policy carries the task state across the context replacement by writing
  and reading the continuation file.} It records the objective, evidence, test status, and next step in
  \texttt{.agent\_memory/CONTINUATION.md} before the replacement, reads the file back after it, and
  completes the task with hidden tests passing (return 1.00). The patch card is an excerpt. The red
  divider marks the replacement.}
  \label{fig:memory-case-study}
\end{figure}

\paragraph{The trained policy writes, reads, and uses memory across a context boundary.}
A single trajectory makes the behavior concrete. We follow one Coding episode from a joint CAMG-RL training run. The task repairs
response logging in a Scrapy codebase. Figure~\ref{fig:memory-case-study} shows the
trajectory terminating at step 30 within Coding's 40-step action budget: the policy creates a
reproducer, patches the missing scheduler method, writes a continuation file before context
replacement, reads it afterward, verifies the fix, and submits.
Appendix~\ref{app:memory-trajectories} presents one complete trajectory from each environment.

%% file: tables/native_heldout_result_table.tex
\begin{table}[!ht]
\centering
\scriptsize
\setlength{\tabcolsep}{2.6pt}
\renewcommand{\arraystretch}{1.08}
\caption{\textbf{CAMG-RL achieves the highest average success rate across the CAMG test set.} We compare
CAMG-RL with training-free memory controls and learned-memory baselines.}
\label{tab:native-heldout}
\begin{tabularx}{\linewidth}{@{}L{0.31\linewidth}*{5}{>{\centering\arraybackslash}X}@{}}
\toprule
& \multicolumn{5}{c}{\textbf{Success rate, \%}} \\
\cmidrule(lr){2-6}
\textbf{Method} & \textbf{Shop} & \textbf{Coding} &
\shortstack{\textbf{Deep}\\\textbf{Research}} &
\shortstack{\textbf{Auto}\\\textbf{Research}} & \textbf{Average} \\
\midrule
\multicolumn{6}{l}{\textit{Frozen and training-free controls}} \\
Qwen3.5-4B
  & 11.6 & 10.9 & 36.7 & 10.2 & 17.4 \\
Mem0 \citep{chhikara2025mem0}
  & 15.6 & 14.8 & 37.5 & 10.9 & 19.7 \\
Letta Code \citep{letta2026lettacode}
  & 12.6 & 14.8 & 32.0 & 11.7 & 17.8 \\
\midrule
\multicolumn{6}{l}{\textit{Learned-memory methods}} \\
CompactionRL \citep{li2026compactionrl}
  & \textbf{97.5} & 19.5 & 50.8 & 24.2 & 48.0 \\
AgeMem \citep{yu2026agenticmemory}
  & 16.7 & \textbf{26.6} & 28.1 & 29.7 & 25.3 \\
\midrule
\textbf{CAMG-RL}
  & 97.1 & \textbf{26.6} & \textbf{55.5} & \textbf{38.3} & \textbf{54.4} \\
\bottomrule
\end{tabularx}
\end{table}

%% file: tables/matched_external_method_comparison_table.tex
\begin{table}[!ht]
\centering
\scriptsize
\setlength{\tabcolsep}{2.4pt}
\renewcommand{\arraystretch}{1.08}
\caption{\textbf{CAMG-RL-4B, trained with file-based memory, matches the 35B-A3B member of its own
family on both benchmarks, and CAMG-RL-9B is competitive with the frozen 122B-A10B.} We compare both
with frozen Qwen3.5 models from 4B to 397B-A17B on SWE-bench Verified and MLE-bench
Lite. We report issue-resolution rate on SWE-bench, Any Medal rate on MLE-bench, and their unweighted
mean as Average.}
\label{tab:matched-external-methods}
\begin{tabularx}{\linewidth}{@{}L{0.39\linewidth}*{3}{>{\centering\arraybackslash}X}@{}}
\toprule
& \multicolumn{3}{c}{\textbf{Success rate, \%}} \\
\cmidrule(lr){2-4}
\textbf{Method} & \shortstack{\textbf{SWE-bench}\\\textbf{Verified}} &
\shortstack{\textbf{MLE-bench}\\\textbf{Lite}} & \textbf{Average} \\
\midrule
Qwen3.5-4B & 7.6 & 0.0 & 3.8 \\
Qwen3.5-9B & 14.8 & 4.5 & 9.7 \\
Qwen3.5-27B & 17.4 & 4.5 & 11.0 \\
Qwen3.5-35B-A3B & 15.6 & 4.5 & 10.1 \\
Qwen3.5-122B-A10B & 22.0 & 9.1 & 15.6 \\
Qwen3.5-397B-A17B & \textbf{34.4} & \textbf{13.6} & \textbf{24.0} \\
\midrule
\textbf{CAMG-RL-4B} & 15.8 & 4.5 & 10.2 \\
\textbf{CAMG-RL-9B} & 27.6 & 9.1 & 18.4 \\
\bottomrule
\end{tabularx}
\end{table}

%% file: sections/discussion.tex
\section{Conclusion}
\label{sec:conclusion}

In this paper, we studied whether reinforcement learning can elicit reusable long-horizon memory
from ordinary file operations. First, we introduced CAMG, a suite
of four agentic-RL environments spanning Shop, Coding, DeepResearch, and AutoResearch, where task
execution depends on information acquired before a context boundary. Alongside each environment's native task interface, CAMG provides a common episode-persistent
workspace and ordinary coding-agent filesystem actions, letting one policy create, revise, search,
and reuse files as memory.
We also introduced \camgrl, which trains a single policy jointly across all four environments
with fully asynchronous PPO, learning this behavior directly from downstream task reward.

Trained this way, \mbox{\camgrl-4B} achieves the highest average success rate across CAMG's four
test environments, ahead of training-free memory systems, learned compaction, and tool-based memory,
and the two trained models remain competitive with Qwen3.5-35B-A3B and Qwen3.5-122B-A10B, respectively,
on SWE-bench Verified and MLE-bench Lite.
By measuring how surprising different memory actions are to the base model before training and
tracking how it learns to use memory, we showed that this advantage comes from fully exploiting
and channeling the pre-trained preference for file operations. Finally, we demonstrated the
learned behavior in a case study: the trained policy writes, reads, and uses memory across a
context boundary.

\paragraph{Limitations.}
The present study limits memory to one episode and one user and trains backbones only up to 9B.
Extending memory across episodes and users, and training larger backbones, are natural
directions for future work.

%% file: sections/statements.tex
\subsection*{AI use statement}

In this work, we use AI for implementing methods, reformatting datasets, and
generating synthetic datasets. We have not used AI for formulating mathematical
claims, providing critical ingredients for proving mathematical claims,
assisting in the writing of
proofs, refining hypotheses, developing theoretical models or conceptual
frameworks, designing research methodology or experiments, assisting with
translation, supporting qualitative and thematic data analysis, or interpreting
results, and the remaining required disclosure tasks are not applicable to this work.
Additionally, we use AI for creating and modifying scientific
figures, creating and editing software code, drafting and editing parts of the
paper, summarizing existing literature, sourcing information, identifying
relevant literature, and suggesting the paper's structure. We have reviewed all
AI-assisted work. We checked mathematical arguments, verified citations against
their sources, tested AI-assisted code through the reported workflows, and
reviewed AI-assisted figures and manuscript revisions. We take responsibility for the final
content of this work, including text, claims or artifacts produced with the aid
of AI.

\subsection*{Ethics statement}

This work does not involve private data or raise ethical concerns.

\subsection*{Reproducibility statement}

Section~\ref{sec:preliminaries} defines the multi-turn POMDP and its complete-response policy-gradient
setting. Section~\ref{sec:camg} instantiates that setting through four long-horizon environments with a
common workspace and context-replacement procedure. Section~\ref{sec:camg-rl} then uses this interface to
define CAMG-RL and its fully asynchronous implementation. Section~\ref{sec:experimental-setup} specifies
joint training and the matched evaluations of the resulting checkpoint. Section~\ref{sec:results}
presents the corresponding comparisons and a qualitative Coding trajectory from joint training.

The appendix follows the same dependency order. Appendix~\ref{app:environments} first describes environment
construction, then the common interface---including its persistent workspace and context transitions---and
task examples. Appendix~\ref{app:learning-systems}
specifies the policy trajectory, credit assignment, asynchronous implementation, training configuration,
and ablation settings. Appendix~\ref{app:evaluation-protocols}
defines CAMG evaluation, failure handling, CAMG-RL ablations, the memory-content intervention,
interface adaptation, and external evaluation.
Appendix~\ref{app:extended-related-work}
reviews additional literature related to the three main-text themes.

We document the datasets and task packages used to construct each environment, the PPO configuration,
the information stored for each episode, and the experiments that produced the reported results. The final
release will include the code needed to reproduce those results.

%% file: sections/appendix.tex
\section*{\centering\LARGE Appendix}

\etocsetnexttocdepth{subsection}
\etocsetlocaltop{part}
\etocsettocstyle{\subsection*{Appendix Contents}}{}
\localtableofcontents

\section{CAMG Environment Details}
\label{app:environments}
\input{sections/appendix_environments}
\input{sections/appendix_interface}
\input{sections/appendix_task_examples}
\input{sections/appendix_memory_trajectories}

\section{CAMG-RL Training Details}
\label{app:learning-systems}
\input{sections/appendix_learning}

\section{Evaluation Details}
\label{app:evaluation-protocols}
\input{sections/appendix_evaluation}

\section{Additional Results}
\label{app:additional-results}
\input{sections/appendix_additional_results}

\input{sections/related_extended.tex}

%% file: sections/appendix_environments.tex
\subsection{Environment Construction}
\label{app:training-envs}

CAMG contains four environments built around long-horizon tasks in which information obtained early can
matter after the active context is replaced. Shop composes new multi-session tasks from WebShop products.
Coding, DeepResearch, and AutoResearch select executable tasks from SWE-smith, LiteResearcher, and public
OpenMLE task packages, respectively. Each environment preserves its domain interaction and definition of task
success. CAMG adds a bounded active context and an episode-private workspace that persists across context
replacement.

\subsubsection{Shop}
\label{app:webshop}

Shop turns six WebShop purchases into one episode \citep{yao2022webshop}. A later request can depend on a
preference, constraint, or clarification established in an earlier session, so the policy must retain the
relevant customer state across browser resets. The interaction itself remains WebShop's native
search--browse--purchase sequence.

We first verify the product catalog, attribute file, price table, and Lucene index. CAMG uses products with
a unique normalized title, first-page search reachability, and matching ASIN and price fields in the purchase
record. The task generator then combines verified products into
six purchase sessions with one cross-session dependency. It fixes the customer, request sequence, candidate
order, dependency branch, and total budget so that exactly one sequence of six purchases is valid. Product
titles are visible to the policy, while the correct product IDs remain hidden and the environment checks
each purchase. Paired branches
preserve later prompts and candidate order while changing the intended target.
Table~\ref{tab:shop-synthetic-families} lists the eight task families.

\input{tables/shop_synthesis_family_table}
\FloatBarrier

A wrong purchase ends the episode. After each correct purchase, the policy updates
\texttt{.agent\_memory/CONTINUATION.md} through the shared procedure in
Appendix~\ref{app:state-action-contract}. Once the environment confirms the write, it clears the browser,
cart, page trace, and active session transcript before presenting the next request. The customer identity,
cumulative budget, purchase history, and episode workspace persist. CAMG Shop contains 720 verified
products: 576 are used for training and 144 for testing. The test products generate 1,746 episodes in
873 counterfactual pairs. The final evaluation set contains 128 episodes from 64 complete pairs.

\subsubsection{Coding}
\label{app:swesmith}

One Coding episode asks the policy to resolve a SWE-smith issue in a specified repository revision
\citep{yang2025swesmith}. Diagnosis, code edits, and test results can span several context windows. The
repository and episode workspace therefore persist when the active messages are replaced, allowing later
actions to reuse earlier evidence. The policy signals a terminal submission with one
\texttt{shell\_command} that prints the fixed sentinel
\texttt{COMPLETE\_TASK\_AND\_SUBMIT\_FINAL\_OUTPUT}, following the mini-swe-agent convention
\citep{yang2024sweagent}. A terminal patch succeeds only when it passes the hidden grader.

Each included task has a complete issue statement and a working hidden grader. Its specified repository
revision and container can be reconstructed, and the reference repair passes the declared tests in that
environment. The resulting collection contains 11,218 executable tasks from 93 repositories, admitted from the
full SWE-smith corpus of 59,136 task instances across 223 repositories. The fixed experimental split
uses 3,768 tasks from 72 repositories for training. The other 7,450 tasks come from 21 disjoint
repositories. We use 933 tasks from 10 repositories as the evaluation pool and sample 128
proportionally across them. The executed training schedule samples from a runtime-health-filtered
panel of 3,792 tasks from 73 repositories: the split's training set plus 24 tasks from one
additional repository, which is disjoint from the evaluation pool and from SWE-bench Verified.

\subsubsection{DeepResearch}
\label{app:literesearcher}

One DeepResearch episode asks the policy to answer a question by alternating between Search and paginated
Visit calls before one terminal Answer \citep{literesearcher2025}. A frozen Kimi-K2.6 semantic judge, shared
by every evaluated method, determines whether the final answer is accepted. Evidence collected early may be
absent from the active context when the answer is produced. The persistent workspace can therefore retain
candidate facts, quotations, page locations, and unresolved conflicts for later use.

We combine two LiteResearcher releases. Each included task has a nonempty question, a search-capable prompt,
one or more valid answer aliases, and corresponding Search and Visit entries.
The resulting collection contains 26,597 tasks: 10,398 from the first release and 16,199 from the second.
Each executable task is paired with its corresponding Search index and paginated Visit corpus. Exact and near-duplicate questions
are kept on the same side of the split, producing 21,278 training tasks and
5,319 test tasks. The final evaluation set contains 128 tasks sampled proportionally across release and
dataset groups.

\subsubsection{AutoResearch}
\label{app:autoresearch}

One AutoResearch episode gives the policy a public OpenMLE task package containing data, a task description,
and a sample submission \citep{yang2026frontisma1}. The policy inspects the data, writes and runs code, and
uses public feedback to improve a candidate solution. Earlier observations, experimental choices, and
intermediate results can be preserved in the workspace across context replacement. A \texttt{submit} action
sends the final submission to the private grader and ends the episode.

CAMG includes packages with nonempty train and test data, a sample submission matching the test rows, a
nonconstant prediction target, and a native metric that can be reproduced under the stated runtime limits.
Each package also has a reference pipeline that completes fitting and prediction, and CAMG records its
public origin and license. Private labels and grader state remain unavailable to the policy. The grader scores the
final submission with the task's native metric. The resulting collection contains 931 tasks from 804
task families.
The fixed split assigns 762 tasks from 664 families to training and holds out 169 tasks from 140 families.
The final evaluation set contains 128 tasks, one from each of 128 families.

%% file: tables/shop_synthesis_family_table.tex
\begin{table}[H]
  \centering
  \small
  \caption{\textbf{Each Shop family requires information to persist across a six-purchase episode.} We
  list the CAMG-synthesized families. Paired variants keep later prompts and candidates fixed while
  changing the relevant earlier state.}
  \label{tab:shop-synthetic-families}
  \begin{tabularx}{\linewidth}{@{}L{0.24\textwidth}X@{}}
    \toprule
    \textbf{Family} & \textbf{Cross-session dependency} \\
    \midrule
    Natural attribute chain & A purchased attribute determines the required attribute in a later category. \\
    Latent preference & Earlier choices establish a preference omitted from later requests. \\
    Recency override & A later preference update replaces an earlier value. \\
    Distractor robustness & The relevant profile must be retrieved among stale and unrelated records. \\
    Compositional recall & Two earlier relations must be composed to identify a later target. \\
    Negative constraint & Earlier exclusions leave one admissible option in later sessions. \\
    Intent clarification & A value elicited in the first session must be reused later. \\
    Selective memory use & Stored preferences matter only when the current request omits the value. \\
    \bottomrule
  \end{tabularx}
\end{table}

%% file: sections/appendix_interface.tex
\subsection{Shared Agent Interface}
\label{app:state-action-contract}

\paragraph{File actions.}
All four environments retain their native task interactions and expose the same single filesystem tool,
following the minimal, shell-only interface of mini-swe-agent \citep{yang2024sweagent}.
\texttt{shell\_command} takes exactly one parameter, a command string, runs every command from the
workspace root, and can create, inspect, search, modify, and execute files. Multi-line file content is
written with one quoted here-document. At each step, the policy produces one complete response. The
environment executes that response once and returns an observation, task reward, and terminal state. Each
response is sampled once, executed once, and trained as one RL step, and it consumes one action slot. A
compound shell command remains one response even when it starts several processes or changes both task
files and memory files.

Each episode receives a fresh private workspace that remains mounted across policy responses and context
replacement. Only regular files below \texttt{.agent\_memory/} count as policy memory. Repository edits,
generated models, temporary files, and caches are not counted as policy memory. When one response changes
both task files and memory files, both changes remain part of that response. The workspace therefore lets the
policy preserve information outside the active context, following the distinction used in language-agent memory
architectures \citep{sumers2024coala}.

\paragraph{Context replacement.}
The environment decides when context pressure or a declared subepisode boundary triggers a
replacement and announces the boundary with a fixed reminder: spend the next response creating
or updating \texttt{.agent\_memory/CONTINUATION.md} through one ordinary filesystem action,
within the configured 8,192-byte limit, following the command shape shown in the reminder.
Complying is the policy's choice, and the environment never writes or clears the continuation
file itself. The requested write consumes one action slot and is sampled and trained like every
other policy response. A response that is not a valid checkpoint write is still executed and
charged. The active context is retained, and a bounded retry notice naming the failure reason
re-requests the checkpoint, so an episode can also end through its task's terminal conditions
with no replacement ever occurring. The environment replaces the active message history only
after it verifies that the policy's own action changed exactly this file into a nonempty
regular file.

After a valid write, the environment replaces the active message history with the fixed task and tool instructions
plus a neutral instruction naming the continuation path. It does not insert the file contents into the new
context. The workspace remains mounted. The policy can recover selected information through a later ordinary
file read. It may also maintain other files under \texttt{.agent\_memory/} and refer to them from the
continuation file.

\paragraph{Execution isolation.}
\label{app:safety-boundary}
Each episode runs in a network-isolated Linux sandbox with a separate writable root and fixed limits on
processes, execution time, output size, file count, and writable bytes. Public task inputs are mounted
read-only. Private credentials and grader state remain outside the sandbox.
Episode termination removes the writable workspace and processes owned by that episode.

%% file: sections/appendix_task_examples.tex
\subsection{Task Examples}
\label{app:task-examples}

\begin{figure}[H]
  \centering
  \includegraphics[width=\linewidth]{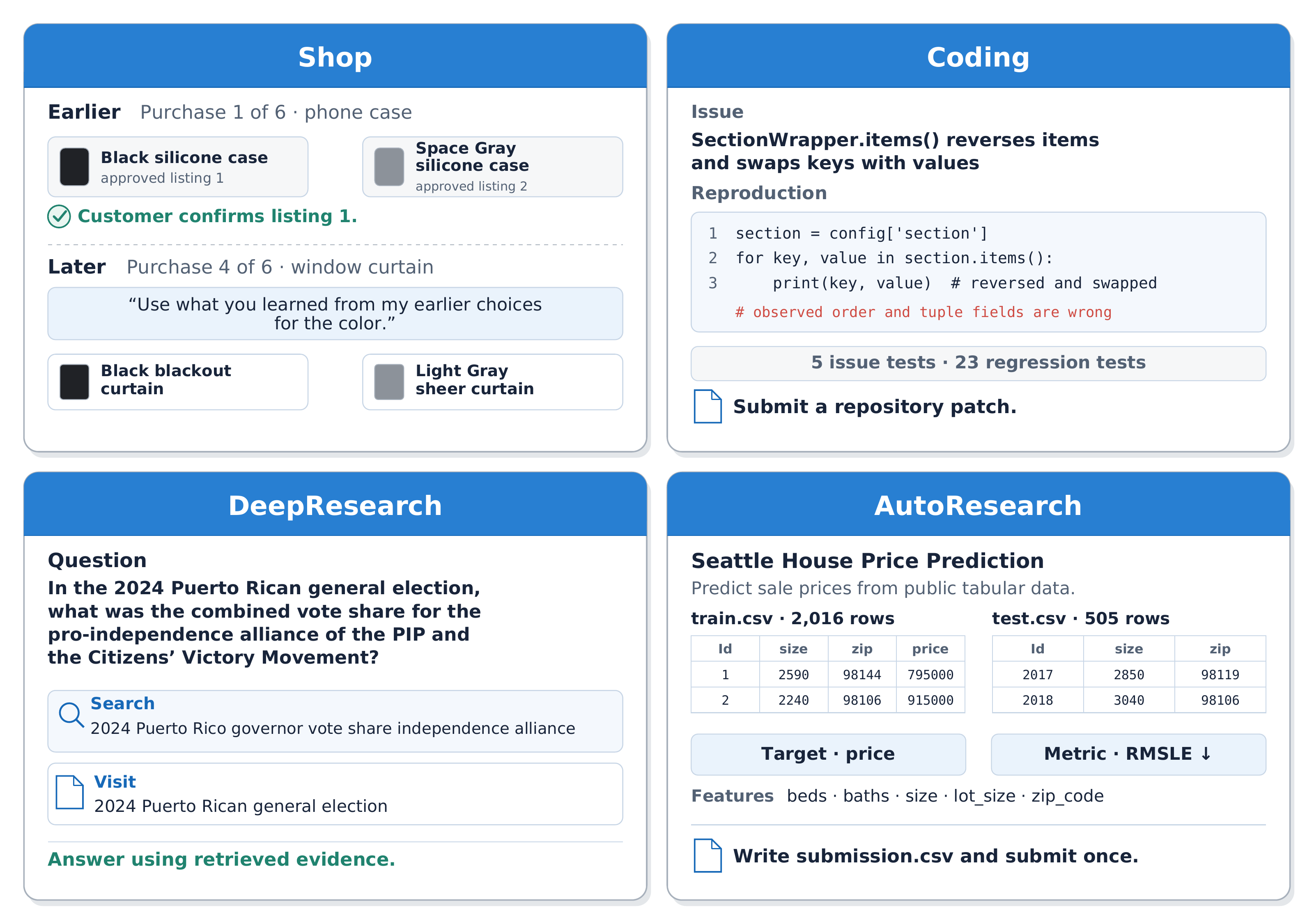}
  \caption{\textbf{Long-horizon CAMG tasks run under a bounded context window.} Each panel summarizes one task input and the interaction available to the policy.}
  \label{fig:camg-task-examples}
\end{figure}

Figure~\ref{fig:camg-task-examples} gives a compact view of one task from each environment. The listings
that follow show the corresponding task inputs and interactions in detail.

\subsubsection{Shop}

This example asks the policy to reuse a color preference established in the first purchase during a later
shopping session.

\begin{minipage}{\linewidth}
\casetitle{Six-session preference task}{Shop}
\begin{lstlisting}[style=taskExcerpt]
[TASK INPUT]
Task ID: amglp.train.color.black_gray.2.e0.063376ff8c661234.t.4cef707f7988d219
Episode: six related shopping sessions

Purchase 1 of 6 - phone case
Option 1: Black silicone case
Option 2: Space Gray silicone case
Feedback: Customer confirms listing 1.

Purchase 4 of 6 - window curtain
Request: Use what you learned from my earlier choices for the color.
Option 1: Black blackout curtain
Option 2: Light Gray sheer curtain

[NATIVE INTERACTION]
search[keywords] -> click[ASIN] -> inspect product/options -> click[Buy Now]

[SHARED FILE INTERFACE]
shell_command: create, inspect, search, modify, or execute files
from the workspace root, with multi-line writes through one quoted
here-document
\end{lstlisting}
\end{minipage}

\subsubsection{Coding}

This example pairs a concrete \texttt{iniconfig} issue with the tests used to verify a repair.

\casetitle{Repository issue and tests}{Coding}
\begin{lstlisting}[style=taskExcerpt]
[TASK INPUT]
Instance: pytest-dev__iniconfig.16793ead.combine_file__mntnwwxj
Repository: pytest-dev/iniconfig

Issue:
SectionWrapper.items() reverses items and swaps keys with values.

Minimal reproduction:
section = config['section']
for key, value in section.items():
    print(key, value)  # current output is reversed and swapped

[VERIFICATION]
Issue tests: 5
Regression tests: 23
Goal: submit a repository patch that passes all hidden tests.

[SHARED FILE INTERFACE]
shell_command operates on the executable repository and policy-authored
notes throughout the episode, from the workspace root.
\end{lstlisting}

\subsubsection{DeepResearch}

This example asks the policy to answer a question using evidence retrieved through Search and Visit.

\casetitle{Question with Search and Visit}{DeepResearch}
\begin{lstlisting}[style=taskExcerpt]
[TASK INPUT]
Example: LiteResearcher task 1521 from the first release

Question:
In the 2024 Puerto Rican general election, what was the combined vote share
for the pro-independence alliance of the Puerto Rican Independence Party and
the Citizens' Victory Movement, marking the first time since 1968 that the
pro-independence movement surpassed one of the two major parties in the
governor's race?

[NATIVE INTERACTION]
search[2024 Puerto Rico governor vote share independence alliance]
visit[policy-visible URL returned by Search]
answer[response supported by retrieved evidence]

[SHARED FILE INTERFACE]
Files may retain document locations, quotations, candidate facts, conflicts,
and open questions during repeated search and reading.
\end{lstlisting}

\subsubsection{AutoResearch}

This example asks the policy to develop a tabular model, produce \texttt{submission.csv}, and submit it once
to the private grader.

\casetitle{Public task package}{AutoResearch}
\begin{lstlisting}[style=taskExcerpt]
[TASK INPUT]
Task ID: house-price-prediction-seattle@1
Title: Seattle House Price Prediction
Metric: RMSLE (lower is better)
Budget: 30 policy actions

Public files:
train.csv - 2,016 rows, target column: price
test.csv  -   505 rows
Features include beds, baths, size, lot_size, and zip_code.

Example public rows:
Id,size,zip,price
1,2590,98144,795000
2,2240,98106,915000

Goal:
Inspect the data, write and revise code, run local validation,
create submission.csv, and invoke submit once when ready.
Submit sends the final file to the private grader.
\end{lstlisting}

%% file: sections/appendix_memory_trajectories.tex
\subsection{Filesystem-Memory Trajectories across a Context Boundary}
\label{app:memory-trajectories}

One successful trajectory per environment appears below. The Shop policy carries confirmed purchase
attributes in \path{.agent_memory/notes.md} across six session boundaries and completes all six
purchases, the DeepResearch and Coding policies store evidence and applied fixes in
\path{.agent_memory/CONTINUATION.md} and reach a correct final answer and passing hidden tests after
the context replacement, and the AutoResearch policy persists its pipeline draft as \path{train.py},
repairs a holdout-split bug in that draft after the replacement, and submits a valid prediction file.
In each case the policy records task state in a workspace file, retrieves it after a context
replacement, and continues with native task actions that reach the environment's success criterion.
The listings that follow place each file-memory write and read inside its surrounding policy action
sequence, while the source evidence records the exact file path, write/read orders, and native
outcome.

\subsubsection{Shop}

This trajectory tracks a cross-session pairing dependency across six shopping sessions. Each session
follows the same search--click--note--buy loop. After confirming the cake flavor in session~1, the
policy writes the attribute to \path{.agent_memory/notes.md}, then writes a continuation checkpoint at
the session boundary and reads it back at the start of every later session.

\casetitle{Cross-session preference store and reuse}{Shop}
\begin{lstlisting}[style=taskExcerpt]
[TASK INPUT]
Six-session shopping task with one cross-session dependency and a
$410 total budget: buy one product per session in order (session 1
- cake base, session 2 - frosting, session 3 - food coloring, session
4 - decorating sprinkles, session 5 - cake topper, session 6 -
cookies); a later request can depend on a preference established in
an earlier session.

[NATIVE INTERACTION]  session 1, turns 1-2
turn 1  search["Betty Crocker Super Moist Triple Chocolate Fudge Cake Mix, 15.25 oz, 2 pk"]
turn 2  click[B00KQ0S8Q2]

[FILE-MEMORY WRITE]  turn 4, session 1 (turn 3 fails without .agent_memory)
shell_command: mkdir -p .agent_memory && cat > .agent_memory/notes.md << 'EOF'
+Confirmed cake flavor: chocolate
EOF

[NATIVE INTERACTION]  session 1, turn 5
turn 5  click[Buy Now]  -> correct purchase (chocolate)

[CONTEXT REPLACEMENT]  turn 6, session boundary
shell_command: mkdir -p .agent_memory && cat > .agent_memory/CONTINUATION.md <<'EOF'
objective: Session 2 of 6 - select confection/decor. Total spent: $100.00. Remaining budget: $310.00.
decisive_evidence: Session 1 bought: Betty Crocker Super Moist Triple Chocolate Fudge Cake Mix, 15.25 oz, 2 pk ($100.00). Note file: .agent_memory/notes.md contains "+Confirmed cake flavor: chocolate".
workspace_paths: .agent_memory/notes.md (read before Session 2), .agent_memory/CONTINUATION.md (this checkpoint)
next_action: Read .agent_memory/notes.md, then search Session 2 approved product title verbatim
EOF

[FILE-MEMORY READ]  turn 7
shell_command: cat .agent_memory/CONTINUATION.md

[NATIVE INTERACTION]  session 2, turns 8-12
turn 8  search["Duncan Hines Whipped Cream Cheese Frosting, 14 OZ"]
turn 10 click[B00L15AOSG]

[FILE-MEMORY WRITE]  turn 11, session 2
shell_command: cat > .agent_memory/frosting_note.md << 'EOF'
+Confirmed frosting flavor: cream cheese
EOF

[NATIVE INTERACTION]  session 2, turn 12
turn 12 click[Buy Now]  -> correct purchase (cream cheese)

[CONTEXT REPLACEMENT + FILE-MEMORY READ]  turns 13-14, session boundary
shell_command: cat > .agent_memory/CONTINUATION.md <<'EOF'
objective: Session 3 of 6 - select cake base. Previous frosting (Session 2) is Duncan Hines Whipped Cream Cheese Frosting, 14 OZ with cream cheese flavor. Determine cake flavor from Session 1 note.
decisive_evidence: Session 2 purchase confirmed: Duncan Hines Whipped Cream Cheese Frosting, 14 OZ (ASIN: B00L15AOSG). Note file: .agent_memory/frosting_note.md contains "Confirmed frosting flavor: cream cheese".
workspace_paths: .agent_memory/CONTINUATION.md, .agent_memory/frosting_note.md, .agent_memory/session1_note.md
next_action: Read .agent_memory/session1_note.md to get Session 1 cake flavor, derive Session 2 cake flavor per pairing table, then search Session 2 approved card title verbatim.
EOF
shell_command: cat .agent_memory/CONTINUATION.md

[NATIVE INTERACTION + FILE-MEMORY WRITE]  sessions 3-5
Sessions 3-5 repeat the cycle: food-coloring note "+Confirmed color: blue",
sprinkles note "+Confirmed listed accent color: silver", cake-topper note
"+Confirmed listed occasion: wedding", each followed by a boundary
checkpoint naming the confirmed attribute and the pairing rule it enables.

[CONTEXT REPLACEMENT + FILE-MEMORY READ]  turns 31-32, session 6 boundary
shell_command: cat > .agent_memory/CONTINUATION.md <<'EOF'
objective: Session 6 of 6 - select dessert: previous cake topper (SOIMISS) has occasion = wedding; customer table requires silver accent -> wedding occasion -> choose VALICLUD 2Pcs Flower Birthday Cake Topper ...
decisive_evidence: Session 5 purchase confirmed: SOIMISS 10 Pcs Heart Cake Topper ... (ASIN: B09PQJ4GMT). Previous sprinkles (Session 4) accent color: silver -> per customer table: silver accent -> choose wedding occasion cake topper.
workspace_paths: .agent_memory/CONTINUATION.md, .agent_memory/sparkles_note.md, .agent_memory/session1_note.md, .agent_memory/frosting_note.md, .agent_memory/session5_note.md
next_action: Search for the VALICLUD cake topper verbatim, open matching result with ASIN, write note with +Confirmed listed occasion: wedding, then click Buy Now
EOF
shell_command: cat .agent_memory/CONTINUATION.md

[NATIVE INTERACTION]  session 6, turns 33-35
turn 33 search["Scott's Cakes Pineapple Shortbread Cookies in a 1 Pound Plastic Deli Container"]
turn 34 click[B00UUN4YC2]
turn 35 click[Buy Now]  -> correct purchase

[NATIVE OUTCOME]
Six of six purchases correct (turns 5, 12, 18, 24, 30, 35).
Episode return 1.00.
\end{lstlisting}

\subsubsection{DeepResearch}

This trajectory accumulates evidence for a factual question over 23 actions. The policy narrows the
search to the Fr.\ John Murphy Monument in Tullow, County Carlow, but the town square's name is still
unknown when a context boundary arrives at action~9, mid-research. The policy writes the objective,
five collected evidence items, and the next actions to \path{.agent_memory/CONTINUATION.md}, reads the
checkpoint back at action~10, and continues the search for another twelve actions, absorbing four
out-of-corpus visit penalties, before producing the final answer.

\casetitle{Evidence checkpoint across a mid-research context replacement}{DeepResearch}
\begin{lstlisting}[style=taskExcerpt]
[TASK INPUT]
Question: A statue unveiled in 1905, made of white Sicilian marble
and standing on a limestone pedestal, commemorates the execution of a
Catholic priest who became a rebel leader during the 1798 Irish
Rebellion and played a key military role in engagements such as the
Battle of Vinegar Hill. This statue is located in a town square that
serves as a central bus stop in a County Carlow town and lies at the
intersection of the N81 and R725 roads. In which square is this statue
located?

[NATIVE INTERACTION]  actions 1-8
action 1  search["statue of Catholic priest rebel leader 1798 Irish
           Rebellion Battle of Vinegar Hill white Sicilian marble
           limestone pedestal County Carlow town square", ...]
action 4  visit[askaboutireland.ie/.../fr.-murphy-monument]
           (Fr. John Murphy Monument in Tullow, County Carlow: white
           Sicilian marble, limestone pedestal, unveiled August 1905,
           executed in Tullow July 1798)
action 5  search["name of square in Tullow County Carlow where
           Fr. Murphy Monument is located", ...]
action 6  visit[en.wikipedia.org/wiki/Monument_Square]
           (disambiguation page, unrelated)
actions 7-8  searches over the N81/R725 intersection; the monument
           page says only "the Square in Tullow" without naming it

[CONTEXT REPLACEMENT]  action 9
shell_command: cat > .agent_memory/CONTINUATION.md <<'EOF'
Objective: Identify the name of the square in County Carlow town where
a statue commemorates the execution of a Catholic priest who became a
rebel leader in the 1798 Irish Rebellion and played a key military role
in the Battle of Vinegar Hill. The statue is made of white Sicilian
marble, stands on a limestone pedestal, was unveiled in 1905, and is
located at the intersection of N81 and R725 roads.

Key Evidence:
1. Fr. John Murphy Monument in Tullow, County Carlow - Statue of Fr.
   John Murphy (Catholic priest, rebel leader in 1798 Irish Rebellion,
   executed in Tullow in July 1798)
2. Statue details: White Sicilian marble, limestone pedestal, unveiled
   in August 1905, designed by Mr. Walsh of Carlow
3. Location: "the Square in Tullow" - situated in Tullow town square
4. Fr. John Murphy took prominent part in 1798 insurrection, captured,
   tortured and executed in Tullow on July 2nd, 1798
5. Source: https://www.askaboutireland.ie/reading-room/... (Fr. Murphy
   Monument page)
[...]
Next actions:
1. Search more specifically for the name of the square in Tullow
   where the Fr. Murphy Monument is located
2. If found, confirm the square name matches all criteria
   (intersection of N81 and R725, central bus stop)
3. If evidence is sufficient, answer the question with the square name
EOF

[FILE-MEMORY READ]  action 10
shell_command: cat .agent_memory/CONTINUATION.md

[NATIVE INTERACTION]  actions 11-22
action 11 search["name of the square in Tullow where Fr. John Murphy
           monument is located", ...]
action 12 visit[askaboutireland.ie/.../fr.-murphy-monument]
           ("The Statue is situated in the Square in Tullow", still
           no specific name)
action 13 search["main square Tullow name", "market square Tullow
           name", "Tullow Bus Station square name", ...]
action 14 visit[en.wikipedia.org/wiki/Eldon_Square]  -> outside the
           released corpus (invalid-visit penalty -0.02)
action 16 visit[en.wikipedia.org/wiki/The_Square]  -> outside the
           released corpus (-0.02)
actions 15, 17-21  further searches; two more out-of-corpus visits
           (-0.02 each)
action 22 visit[en.wikipedia.org/wiki/Tullow]

[NATIVE OUTCOME]  action 23
Based on the Wikipedia page for Tullow: "There is a statue of Father
John Murphy, one of the leaders of the 1798 Rebellion, who was captured
near Tullow and executed in the Market Square on 2 July."
answer[Market Square]  -> accepted under the fixed semantic judge.
Episode return 0.92 (answer reward 1.00 minus four -0.02
out-of-corpus visit penalties).
\end{lstlisting}

\subsubsection{Coding}

This trajectory repairs a repository issue and terminates at round 30 within Coding's 40-round budget. The policy materializes the issue's
reproduction spider as \path{test_spider.py}, confirms the scheduler defect on the persisted script,
and applies a scripted fix. A context boundary arrives at round~18, before verification: the
checkpoint records the applied fix, the persisted test script, and the next action. The policy reads
the checkpoint back at round~19, re-runs the persisted script at round~20, and completes verification
with the repository's own test suite.

\casetitle{Checkpoint and persisted test script across a context replacement}{Coding}
\begin{lstlisting}[style=taskExcerpt]
[TASK INPUT]
Instance: scrapy - response logging not working after recent changes.
Repository: scrapy
Goal: submit a repository patch that passes all hidden tests.

[NATIVE INTERACTION + FILE WRITE]  rounds 1-4 (persist the reproduction)
round 1  shell_command: ls
round 2  shell_command: python -c "import scrapy; ... process.start()"
          (AttributeError: 'Scheduler' object has no attribute '_mqpush',
           scrapy/core/scheduler.py line 259)
round 3  shell_command: cat > test_spider.py << 'EOF'
          (the issue's reproduction spider, written as a workspace file)
round 4  shell_command: python test_spider.py 2>&1 | head -100
          (the error is confirmed on the persisted script)

[NATIVE INTERACTION]  rounds 5-12 (locate and fix)
rounds 5-11  enqueue_request calls self._dqpush(request) and then
          self._mqpush(request); grep shows _dqpush exists but _mqpush
          is missing, and self.mqs has pop() but no push method
round 12 shell_command: cat > /tmp/fix_scheduler.py << 'EOF' ... EOF
          && python /tmp/fix_scheduler.py
          (adds the missing _mqpush method after _dqpop in
           scrapy/core/scheduler.py)

[NATIVE INTERACTION]  rounds 13-17 (verify with the persisted script)
round 13 shell_command: python test_spider.py 2>&1 | head -60
          (the AttributeError is gone; only an expected DNS lookup
           failure for example.com remains)
rounds 15-17  read tests/test_crawl.py to see how response logging
          is verified ("Got response 200")

[CONTEXT REPLACEMENT]  round 18
shell_command: cat > .agent_memory/CONTINUATION.md <<'EOF'
objective: Fix response logging broken after recent changes -
AttributeError: '_mqpush' missing in Scheduler

evidence:
1. Fixed: added _mqpush method to scrapy/core/scheduler.py after
   _dqpop method
2. Tested: test_spider.py runs without AttributeError anymore
3. The spider now fails due to DNS lookup failure for example.com
   (expected in test env), not code error
4. No more "AttributeError: 'Scheduler' object has no attribute
   '_mqpush'" error

paths:
- scrapy/core/scheduler.py (modified)
- test_spider.py (test file)
- .agent_memory/CONTINUATION.md (this checkpoint)

next: Run tests to verify the fix and confirm logging works, then
submit with final output
EOF

[FILE-MEMORY READ]  round 19
shell_command: cat .agent_memory/CONTINUATION.md

[NATIVE INTERACTION]  round 20 (the persisted script crosses the boundary)
round 20 shell_command: python test_spider.py
          (the fix holds after the replacement; the DNS failure is
           environmental)

[NATIVE INTERACTION]  rounds 21-29 (verify with the repository suite)
rounds 23-25  shell_command: python -m pytest -v -k log;
          python -m pytest tests/test_scheduler.py -v -k log
round 27 shell_command: python -m pytest tests/test_crawl.py -v
          (response logging visible in the test output)
round 29 The fix for `_mqpush` has been applied and the documentation
          from the continuation note confirms the issue is resolved.
          shell_command: grep -A 3 "_mqpush" scrapy/core/scheduler.py

[NATIVE OUTCOME]
round 30 shell_command: echo COMPLETE_TASK_AND_SUBMIT_FINAL_OUTPUT
Hidden tests pass.
Episode return 1.00.
\end{lstlisting}

\subsubsection{AutoResearch}

This 23-action trajectory addresses an Indonesian learning-style classification task over free-text
study-habit descriptions. The policy authors \path{train.py} through a quoted here-document and
iterates on failing drafts for fifteen actions without a single successful run. A context boundary
at action~16 leaves an unfilled checkpoint, so the policy reads the workspace back, diagnoses a
holdout-split bug directly in its persisted draft, and repairs the pipeline. The repaired script
completes the first successful run at action~21, both one-time milestones fire at once, and the
episode submits a valid prediction file. The listing summarizes the intermediate rewrites and
failed runs.

\casetitle{Workspace-persisted draft diagnosed and repaired after context replacement}{AutoResearch}
\begin{lstlisting}[style=taskExcerpt]
[TASK INPUT]
Task: predict the dominant learning style (Visual, Auditory, or
Kinesthetic) for the test rows in data/test.csv from Indonesian
free-text study-habit descriptions, using at most 30 actions.
Deliverables: train.py that prints a measured validation metric,
and submission.csv matching data/sample_submission.csv row
identifiers.

[NATIVE INTERACTION]  action 1
shell_command: cat data/description.txt; echo "---"; head -5 data/sample_submission.csv; echo "---"; head -5 data/test.csv; echo "---"; head -5 data/train.csv

[FILE WRITE]  action 2 (pipeline authored with one here-document)
shell_command: cat > train.py << 'EOF'
import pandas as pd
import numpy as np
...
# Simple holdout split for validation (10% of train)
val = train.sample frac=0.1, random_state=42
...
X_train_vec = vectorizer.transform(X_train)
...
print(f'validation_macro_f1={_macro_f1:.4f}')
...
EOF

[NATIVE INTERACTION]  actions 3-15 (iterate on failing drafts)
action 3  shell_command: python train.py  (syntax error in the
          validation split)
action 4  rewrite switches to train_test_split, but assigns
          X_train, y_train = split[0], split[1]
          X_val, y_val = split[0], split[1]
          (the same half is reused for train and validation)
action 6  rewrite replaces the nonexistent metric import with a
          plain accuracy computation
actions 8-12  add fit_transform, drop stop_words, and relax
          min_df while the split bug persists
action 14 prose: the error indicates that `y_train` is being
          treated as a 2D array ... rewrite converts the labels
          with .values
action 15 shell_command: python train.py  (still failing, no
          milestone earned before the boundary)

[CONTEXT REPLACEMENT]  action 16
shell_command: mkdir -p .agent_memory && printf '%s
' objective=OBJECTIVE measured_validation_or_failure=VALIDATION_OR_FAILURE conclusion=CONCLUSION code_path=train.py next_action=NEXT_ACTION > .agent_memory/CONTINUATION.md
(the checkpoint is written with the template unfilled)

[FILE-MEMORY READ]  action 17
shell_command: cat .agent_memory/CONTINUATION.md

[NATIVE INTERACTION]  actions 18-23 (recover from the workspace)
action 18 The continuation file is empty. Let me check the current
          state of the workspace to understand what's been done and
          what's needed.
          shell_command: ls -la /workspace/ && ...  (draft and data
          files present, no submission.csv yet)
action 19 Let me check the current state of train.py and the data
          files to understand what's been done:
          shell_command: cat /workspace/train.py && ...
action 20 I can see there's an issue with the train.py script. The
          holdout split logic has a bug - `train_test_split[0]` is
          used twice, so train and validation share one half. Let me
          fix this and run the script to get a proper validation
          metric:
          shell_command: cat > /workspace/train.py << 'EOF'
          ...
          X_train, X_val, y_train, y_val = train_test_split(
              train['text'], train['label'],
              test_size=0.1, random_state=42,
              stratify=train['label'])
          ...
          print(f'validation_accuracy={accuracy:.4f}')
          ...
          EOF
action 21 shell_command: python /workspace/train.py
          (first successful code execution and first measured
          validation metric together: one-time +0.05 milestone and
          one-time +0.10 milestone, validation_accuracy=0.9241)
action 22 prose: the script runs successfully with a validation
          accuracy of 0.9241 and creates submission.csv ...
          shell_command: head -10 /workspace/submission.csv && ...
          (197 lines matching the 196 test rows plus header)

[NATIVE OUTCOME]
action 23 - submit.
Valid submission accepted by the native grader.
Episode return 1.14 (execution milestone 0.05 + validation milestone
0.10 + terminal reward 0.99).
\end{lstlisting}

Across these trajectories, the same filesystem interface supports preference tracking, evidence accumulation, repository debugging, and experiment iteration through policy-authored state that remains available throughout the task.

%% file: sections/appendix_learning.tex
\subsection{Policy Trajectory}
\label{app:camg-rl-procedure}

With CAMG-RL, we train one policy on trajectories that include both native task actions and filesystem
operations. At turn $t$, the policy samples one complete response $a_t=y_{t,1:L_t}$, and the environment
executes it once. Each complete response forms one RL step:
\begin{equation}
  \mathbf{z}_t=(h_t,a_t,\boldsymbol{\ell}^{\,b}_t,r_{d,t},e_t,v_t^b),
  \qquad
  \ell^{b}_{t,j}=\log \pi_{\theta_b}(y_{t,j}\mid h_t,y_{t,<j}).
  \label{eq:action-row}
\end{equation}
Here $h_t$ is the bounded policy-visible context, $\boldsymbol{\ell}^{\,b}_t$ contains the behavior log
probabilities, $r_{d,t}$ is the reward returned after the response, $e_t$ marks episode termination, and
$v_t^b$ identifies the behavior-policy version that generated the response. Because every response enters
the same trajectory, a filesystem response can receive credit from task rewards observed later in the
episode. A write to \texttt{.agent\_memory/CONTINUATION.md} is sampled like any other policy response and
constitutes one RL step. The subsequent context reset only changes the next $h_t$. It contains no sampled
tokens and adds no PPO loss term.

\subsection{Credit Assignment}
\label{app:action-credit}

We use the environment rewards defined in Appendix~\ref{app:training-details}. At turn $t$,
$r_{d,t}$ is the reward returned by the environment after $a_t$. For sampled token $j$, define the
pre-token critic value
\begin{equation}
  V_{t,j}^{-}:=V_{\phi^{-}}(h_t,y_{t,<j}),
  \qquad
  \bar V_t:=V_{t,1}^{-}=V_{\phi^{-}}(h_t)
  \approx \mathbb{E}[G_t\mid H_t^{\mathrm{vis}}=h_t],
  \label{eq:conditional-value}
\end{equation}
where $\phi^{-}$ denotes the critic before that learner update. The actor and critic receive
the same bounded policy-visible context. The critic does not observe hidden native state or the contents
of files that the policy has not read. We compute GAE across response steps:
\begin{align}
  \delta_t &= r_{d,t} + \gamma(1-e_t)\bar V_{t+1}-\bar V_t,
  \label{eq:action-gae} \\
  \widehat A_t &= \delta_t+\gamma\lambda(1-e_t)\widehat A_{t+1}, \\
  G_t &= \widehat A_t+\bar V_t,
\end{align}
where $\bar V_{T+1}=\widehat A_{T+1}=0$ and $\gamma,\lambda\in[0,1]$. We set
$\gamma=\lambda=1$. The learner assigns
$\widehat A_t$ and $G_t$ to every sampled token in response $t$, giving the raw token advantage
\begin{equation}
  A^{\mathrm{raw}}_{t,j}=\widehat A_t.
  \label{eq:token-advantage}
\end{equation}
We center these advantages within each environment over non-padding policy tokens and normalize them
using a scale shared across the learner batch. We then apply inverse environment token-count weights
$N_{\mathrm{tok}}/(D_{\mathrm{batch}}N_d)$ and a final masked standardization over the same batch.
Here $N_{\mathrm{tok}}$ is the total number of non-padding policy tokens, $N_d$ is the count for
environment $d$, and $D_{\mathrm{batch}}$ is the number of environments present. These operations
affect actor advantages only. Critic return targets $G_t$ are unchanged. We use $A_{t,j}$ below to
denote the resulting normalized and weighted advantages.

For each sampled token, the current-to-behavior probability ratio is
\begin{equation}
  \rho_{t,j}(\theta)=
  \exp\!\left[
    \log\pi_\theta(y_{t,j}\mid h_t,y_{t,<j})-\ell^b_{t,j}
  \right].
  \label{eq:token-ratio}
\end{equation}
For optimizer mini-batch $k$, let $\mathcal R_k$ be its responses and
$\mathcal I_k=\{(t,j):t\in\mathcal R_k,1\leq j\leq L_t\}$. With policy clip $\epsilon=0.2$ and value
clip $\epsilon_V=0.5$, define
\begin{equation}
  I_{t,j}=[V_{t,j}^{-}-\epsilon_V,V_{t,j}^{-}+\epsilon_V],
  \qquad
  \widetilde V_{t,j}(\phi)=
  \operatorname{clip}\!\left(V_\phi(h_t,y_{t,<j}),I_{t,j}\right).
\end{equation}
The actor and critic losses are
\begin{align}
  \mathcal{L}_{\pi}^{(k)}(\theta)
  &=-\frac{1}{|\mathcal{I}_k|}\sum_{(t,j)\in\mathcal{I}_k}
  \min\!\left(
    \rho_{t,j}A_{t,j},
    \operatorname{clip}(\rho_{t,j},1-\epsilon,1+\epsilon)A_{t,j}
  \right), \\
  \mathcal{L}_{V}^{(k)}(\phi)
  &=\frac{1}{2|\mathcal{I}_k|}\sum_{(t,j)\in\mathcal{I}_k}
  \max\!\left[
    \bigl(V_\phi(h_t,y_{t,<j})-G_t\bigr)^2,
    \bigl(\widetilde V_{t,j}(\phi)-G_t\bigr)^2
  \right].
  \label{eq:action-ppo}
\end{align}
Every sampled token in a response receives the same normalized response-level advantage, so longer responses
contribute more terms to the actor loss. Later task outcomes can therefore update earlier filesystem
responses, including responses that write or retrieve memory.

\paragraph{No gradient from memory actions.}
For this control in Figure~\ref{fig:camg-rl-ablation-expected}, we keep rewards, returns, critic targets, and critic loss
unchanged but omit from the actor loss every response that accesses a file under
\texttt{.agent\_memory/}, whether or not the operation succeeds. If one response also accesses task files,
we omit the whole response from the actor loss. These responses still change the workspace and count
toward the same action, token, runtime, and failure budgets. They receive no direct actor gradient, although
updates from other responses can still change later memory behavior.

\subsection{Asynchronous Training}
\label{app:fully-async-algorithm}

Episode duration varies substantially across the four environments. Synchronous collection would leave
workers idle while the slowest interactions finish. We therefore let rollout and learning proceed
independently. Rollout workers use the same rollout process for all four environments and append completed
episodes to a persistent queue. The learner repeatedly consumes the earliest-dispatched complete episodes. Each environment
retains its task-specific transitions, context-replacement conditions, rewards, and termination rules. The
learner consumes 64 episodes per parameter update. Within those episodes, each complete policy response
is one action-level training example, and PPO minibatches contain 510 such responses. The episode count and
the response count therefore refer to different levels of the training procedure.

\begin{center}
  \refstepcounter{algorithm}\label{alg:camg-fully-async}
  \setlength{\fboxsep}{6pt}
  \fbox{\begin{minipage}{0.95\textwidth}
  \small
  \textbf{Algorithm~\thealgorithm: Fully asynchronous CAMG-RL training}\\[2pt]
  \begin{tabularx}{\textwidth}{@{}rX@{}}
  \textbf{Input:} & fixed task schedule, four environments, actor, critic, episode
  queue $\mathcal Q$, and learner batch size $N$. \\
  \multicolumn{2}{@{}l}{\textbf{Rollout workers, running continuously:}} \\
  1 & Take the next scheduled episode ID and initialize its task, native state, bounded context, and
  episode-private workspace. \\
  2 & Sample one complete response from the latest actor using the fixed EOS-or-length stopping
  rule. Save its context, sampled tokens, behavior log probabilities, stopping reason, and policy version. \\
  3 & Call $\operatorname{env.step}(a_t)$ and append the returned reward and terminal state. Apply any
  context replacement before sampling the next response. Continue until the environment terminates the
  episode. If the response limit $H_d$ is reached, assign the final outcome to the last sampled response. \\
  4 & Enqueue the completed episode. Infrastructure faults follow the rerun rule in
  Appendix~\ref{app:failure-taxonomy}. \\
  \multicolumn{2}{@{}l}{\textbf{Learner, running independently:}} \\
  5 & Dispatch the next $N$ prompts to the rollout workers, wait until the queue holds $N$ complete
  episodes, and select the $N$ earliest dispatched. Concatenate their responses, and pad tensors with
  masked positions where needed. \\
  6 & Compute action-axis GAE and the token-level PPO losses above. Update the actor and critic, then
  make the updated actor available to rollout workers after each learner update. \\
  7 & Save the identities of the episodes used, their policy versions, the optimization statistics, and
  the checkpoint state. \\
  \textbf{Output:} & one trained CAMG-RL policy.
  \end{tabularx}
  \end{minipage}}
\end{center}

Only complete episodes enter the queue, and each scheduled episode is used in at most one learner update.
Episodes that complete with task failure remain training examples. Infrastructure faults follow the rerun rule in
Appendix~\ref{app:failure-taxonomy}. Because episode durations differ, the set of complete episodes available
at each learner update varies, so the environment mixture varies across learner updates.

A small warmup window and the current dispatch are the only prompts outstanding, so generation stays
close to the learner without any admission cap. Every queued or active response retains the
behavior-policy version that generated it, while only complete episodes enter the learner. For a
response used at learner version $v_k^\ell$, policy-version lag is $\Delta v_t=v_k^\ell-v_t^b$. The
clipped-PPO ratio compares current token probabilities with the stored behavior probabilities.

\paragraph{Efficiency and optimization measurements.}
\label{app:runtime-accounting}
At each policy update, we record three groups of measurements. Optimization health includes policy-version
lag, PPO ratio tails, clip fraction, and effective sample size. Workload composition includes each
environment's episode, policy-response, and token shares together with queue occupancy and waiting time.
Throughput includes completed episodes, training iterations, sampled tokens, policy responses, accelerator-hours,
and time spent in rollout, environment execution, learning, synchronization, and checkpointing. Task
outcomes are reported separately.

\subsection{Training Configuration}
\label{app:training-details}

CAMG-RL jointly trains on the four environments with equal episode allocation. We update all actor and
critic parameters. Table~\ref{tab:joint-training-config} gives the shared training configuration.

\input{tables/fully_async_ppo_reference_config_table}

Let $T$ denote the terminal response and $\widetilde r_t$ the native environment reward.
Table~\ref{tab:family-reward-contract} defines the reward $r_t$ used in action-axis GAE. At the
response limit, the terminal reward is assigned to the final sampled response. Shop retains rewards
from earlier successful purchases.

\input{tables/family_reward_contract_table}

\begin{figure}[H]
  \centering
  \includegraphics[width=0.98\linewidth]{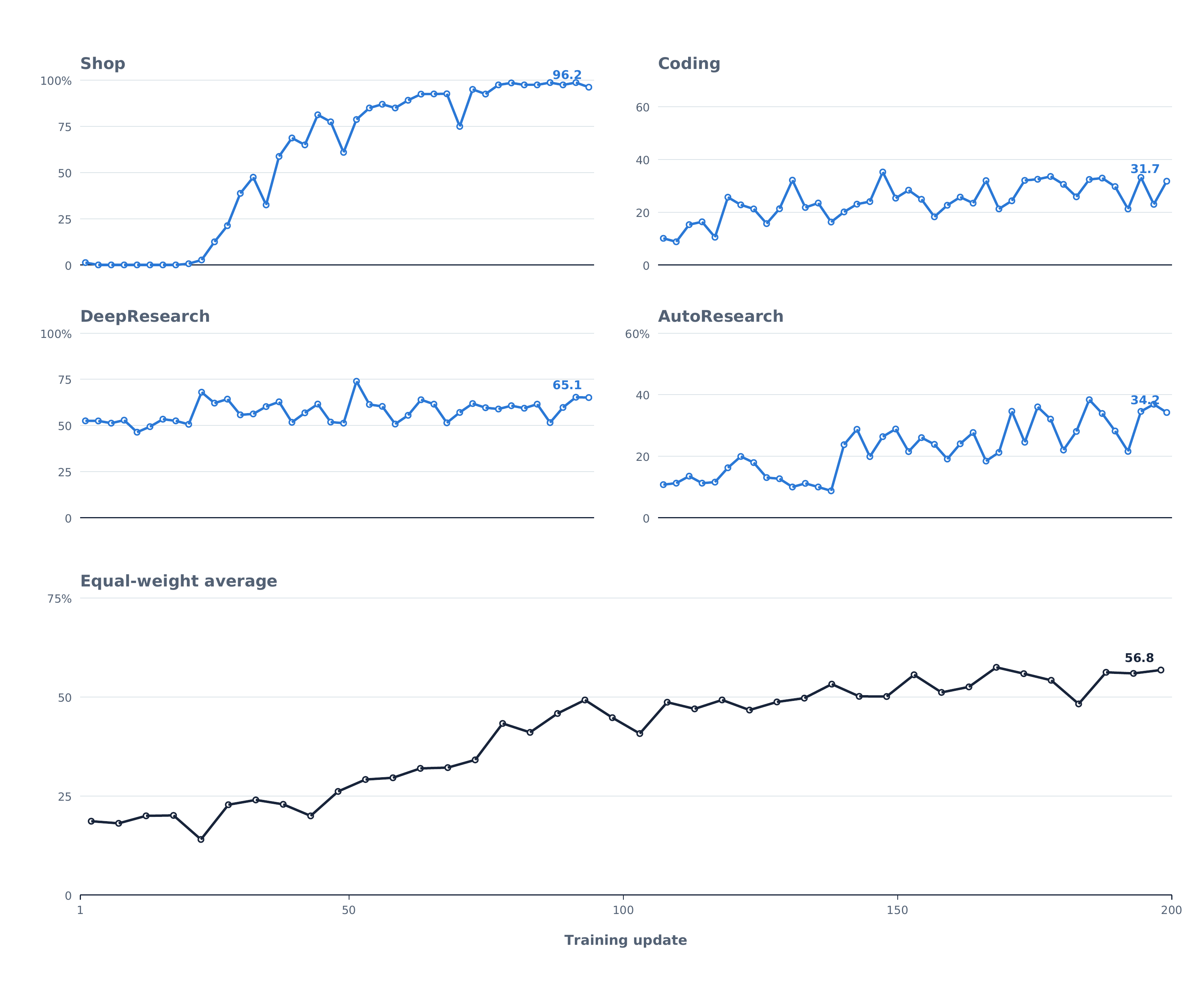}
  \caption{\textbf{All four environments improve during the 200-update joint run.} Each panel shows one
  environment's success rate on training rollouts with temperature-1 sampling and 64 episodes per
  update, and the bottom panel shows the equal-weight average. Lines are non-overlapping five-update
  bucket means, and each panel annotates the final bucket of the checkpoint
  evaluated in Table~\ref{tab:native-heldout}.}
  \label{fig:camg-rl-training-curve}
\end{figure}

\paragraph{Training dynamics.}
Figure~\ref{fig:camg-rl-training-curve} reports the success rate measured on training rollouts
throughout the 200-update joint run. These are temperature-1 rollouts on the training task pools, not
test-set performance, and the update-200 checkpoint is the one evaluated in
Table~\ref{tab:native-heldout}. The plotted lines show non-overlapping five-update bucket means over
the 64 episodes completed in each update, with markers identifying the bucket centers.

%% file: tables/fully_async_ppo_reference_config_table.tex
\begin{table}[H]
  \centering
  \small
  \renewcommand{\arraystretch}{0.98}
  \setlength{\tabcolsep}{4pt}
  \caption{\textbf{CAMG-RL uses shared training settings with per-environment action budgets.}}
  \label{tab:joint-training-config}
  \begin{tabularx}{\linewidth}{@{}p{0.17\linewidth}p{0.37\linewidth}X@{}}
    \toprule
    Group & Hyperparameter & Value \\
    \midrule
    Model & Backbone & Qwen3.5-4B \\
    \midrule
    Schedule & Parameter updates & 200 \\
         & Episodes per update & 64 \\
         & Episode sampling order & Oldest dispatched first \\
         & Episodes per prompt & 1 \\
         & Maximum action steps per episode & Shop 42, Coding 40, DeepResearch 40, AutoResearch 30 \\
    \midrule
    PPO & Discount factor $\gamma$ & 1 \\
        & GAE parameter $\lambda$ & 1 \\
        & Advantage normalization & Masked standardization \\
        & Policy clip & $0.2$ \\
        & Value clip & $0.5$ \\
        & PPO epochs & 1 \\
        & Entropy coefficient & 0 \\
        & KL coefficient & 0 \\
    \midrule
    Optimization & Optimizer & AdamW \\
                 & Policy learning rate & $10^{-6}$ \\
                 & Value-function learning rate & $10^{-5}$ \\
                 & Weight decay & $0.01$ \\
                 & Gradient norm limit & $1.0$ \\
                 & PPO minibatch size & 510 \\
                 & Policy examples per GPU step & 8 \\
                 & Value examples per GPU step & 8 \\
    \midrule
    Sequence length & Prompt token limit & 30,720 \\
                   & Response token limit & 8,192 \\
                   & Observation token limit & 8,192 \\
    \midrule
    Generation & Temperature & 1 \\
               & Nucleus sampling threshold ($p$) & 1 \\
               & Stopping rule & First EOS or response-length limit \\
               & Maximum tokens processed together & 131,072 \\
               & Maximum sequences processed together & 32 \\
    \midrule
    Runtime & Distributed training & FSDP2 \\
            & Training data type & BF16 \\
            & Memory-saving technique & Gradient checkpointing \\
            & Text generation system & SGLang \\
            & Training GPUs & 6 \\
            & Generation GPUs & 2 \\
    \bottomrule
  \end{tabularx}
\end{table}

%% file: tables/family_reward_contract_table.tex
\begin{table}[H]
\centering
\small
\renewcommand{\arraystretch}{1.08}
\caption{\textbf{Task rewards are sparse and success-based across the four environments.}}
\label{tab:family-reward-contract}
\begin{tabularx}{\linewidth}{@{}L{0.18\linewidth}X@{}}
\toprule
\textbf{Environment} & \textbf{Reward} $r_t$ \\
\midrule
Shop & $r_t=\widetilde r_t/7$ when $\widetilde r_t>0$, and $r_t=\widetilde r_t$ otherwise. Correct purchases give $\widetilde r_t=1$ in sessions 1--5 and $\widetilde r_t=2$ in session 6. Invalid actions and incorrect purchases give $\widetilde r_t=-0.01$ \\
Coding & $r_T=\mathbf{1}\{\text{hidden tests pass}\}$ and $r_t=0$ for $t<T$. A context-boundary response whose checkpoint write fails to persist receives $r_t=-0.1$ \\
DeepResearch & $r_T=\mathbf{1}\{\text{semantic judge accepts the answer}\}$ and $r_t=0$ for $t<T$. Invalid actions, including failed checkpoint writes, receive $r_t=-0.02$, and the episode ends at the twelfth invalid action \\
AutoResearch & On a valid submission, $r_T=\min(1,\max(0,s_T)+0.1)$ for direction-normalized private-grader score $s_T$, where the $+0.1$ is a one-time first-submission milestone. Invalid submissions give $r_T=0$, exhausting the action budget without submitting gives $r_T=-0.01$, and exceeding the stated runtime limits gives $r_T=-1$. For $t<T$, $r_t=0$ except two one-time milestones: $+0.05$ for the first successful code execution and $+0.1$ for the first recorded validation metric. Unparseable actions receive $r_t=-0.01$ \\
\bottomrule
\end{tabularx}
\end{table}

%% file: sections/appendix_evaluation.tex
\subsection{CAMG Evaluation}
\label{app:evaluation-details}

CAMG evaluation uses 128 test tasks per environment. The split is disjoint by product for Shop,
repository for Coding, duplicate question group for DeepResearch, and task family for AutoResearch. All
methods use the same 512 tasks, decoding, action budgets, runtimes, interfaces, and
graders. Every high-level memory operation in a training-free control consumes one global action-budget
step. Its internal model, embedding, vector-store, or git subcalls are retained in the cost record but do
not consume additional policy-action steps. Training-free controls begin each task with empty memory, so
memories never carry over across tasks. In the variant evaluated without general memory files in Figure~\ref{fig:camg-rl-ablation-expected}, the terminal CAMG-RL
checkpoint can access \texttt{.agent\_memory/CONTINUATION.md} as its only memory file. Its native task
actions and task files remain unchanged.

\paragraph{CAMG-RL ablations.}
The component comparison uses the same tasks and runtime settings as the main CAMG evaluation.
The policy without memory-action gradients omits every memory-access response from the actor loss while
retaining rewards, returns, critic targets, and critic training. The policy retrained without general memory files uses the bounded continuation
file as its only memory file in training and evaluation. The variant evaluated without general memory files
applies that restriction to the terminal full-memory CAMG-RL policy without additional training. It retains more of Coding, DeepResearch, and AutoResearch than the retrained policy but
collapses Shop to 0\% (Figure~\ref{fig:camg-rl-ablation-expected}). The fully trained policy keeps
attempting to write memory files beyond the continuation file, whereas the policy trained under the
restriction consolidates its memory into that file.

\paragraph{Success rates.}
Let $N=128$ be the number of evaluated episodes in each environment. Shop contains six fixed shopping
sessions per episode, so its success rate is $\sum_{i=1}^{N} c_i/(6N)$, where
$c_i\in\{0,\ldots,6\}$ is the number of completed sessions. Coding is the fraction of episodes whose
submitted patch resolves the issue under hidden tests. DeepResearch is the fraction with an accepted
terminal answer under the fixed semantic judge. AutoResearch is the fraction with a valid submission
whose direction-adjusted private score strictly beats the frozen generic baseline. Invalid or missing
submissions count as failures. Average Success is the unweighted mean of the four rates.

AutoResearch also retains a continuous quality diagnostic. For raw score $z$, frozen generic baseline
score $b$, ideal native score $z^\star$, and metric direction $d\in\{-1,+1\}$, it is
\begin{equation}
  u=\operatorname{clip}\!\left(
    \frac{d(z-b)}{d(z^\star-b)},-1,1
  \right).
  \label{eq:autoresearch-utility}
\end{equation}
Here $d=+1$ for metrics that are maximized and $d=-1$ for metrics that are minimized. Every evaluated
task satisfies $d(z^\star-b)>0$. An invalid or missing submission receives $u=-1$. We report mean
utility and the valid-submission rate as secondary diagnostics.

\paragraph{Training-free action accounting.}
\label{app:action-accounting}
Across the 512 test tasks, Mem0's automatic memory manager consumes 2,778 high-level memory
steps, 5.43 per task, and the same evaluation contains 10,603 native task actions, down from
12,535 for the frozen base. Letta Code issues 110 explicit MemFS actions among 13,399 action
rows.

\paragraph{Paired comparisons.}
Let $s_e^{(k)}$ be the success contribution of episode $e$ for method $k$: completed sessions divided by
six for Shop and a binary success indicator for each other environment. For matched methods $k$ and
$k'$, the task-level difference is
\begin{equation}
  \widehat\Delta_{k,k'}=\frac{1}{|\mathcal E|}
  \sum_{e\in\mathcal E}\left(s_e^{(k)}-s_e^{(k')}\right).
  \label{eq:paired-outcome}
\end{equation}
We assess each paired difference with 10,000 paired bootstrap resamples of task IDs and treat a
difference as significant when its 95\% interval excludes zero.
Resampling is stratified by environment, redrawing each environment's 128 task IDs. The resulting average-success intervals, in percentage points, are $[51.6, 57.2]$
for CAMG-RL, $[45.4, 50.6]$ for CompactionRL, and $[22.7, 27.9]$ for AgeMem. The paired
CAMG-RL$-$CompactionRL difference of $6.4$ points has interval $[3.3, 9.5]$ and the
CAMG-RL$-$AgeMem difference has interval $[26.6, 31.6]$, both excluding zero. Per environment, the
CAMG-RL$-$CompactionRL intervals are $[1.2, 12.8]$ for Coding, $[5.6, 22.8]$ for AutoResearch,
$[-1.4, 11.4]$ for DeepResearch, and $[-1.7, 0.9]$ for Shop, so the advantage is significant on
Coding and AutoResearch, DeepResearch does not separate, and both methods sit at the Shop
ceiling (97.1 versus 97.5).

\label{app:failure-taxonomy}
Invalid actions, failed commands, incorrect outputs, exhausted budgets, and missing submissions count as
failures within the fixed 128-task denominator. An infrastructure fault may be rerun at most twice with
the same task, checkpoint, and runtime settings. If the fault remains unresolved, the corresponding
method--environment cell remains \texttt{N/A} until all 128 assigned tasks have results.

\paragraph{Content intervention.}
A trained-policy test trajectory must have at least one nonempty
policy-created memory file before its first completed context replacement. Immediately after that reset
and before the next policy response, we fork the same post-reset state into three conditions. The intact condition
leaves all memory files unchanged. The blank condition preserves their paths but removes their contents. The
shuffled condition replaces their contents with length-matched content from a different eligible task in the
same environment. Donors follow a fixed within-environment permutation matched by byte length. The
policy, native task state, policy-visible post-reset context, non-memory files, sampling state, action and
token budgets, interface, and runtime remain identical across the three forks. We report
the success-rate change of blank and shuffled memory relative to intact memory for each environment and
their equal-weight average. This intervention tests whether saved content changes later policy behavior while filesystem actions and
memory-file paths remain available in every condition.

\paragraph{Interface-prior probes.}
\label{app:interface-prior-analysis}
We compare CAMG-RL's filesystem actions with the six dedicated tools used by the AgeMem baseline using three diagnostics.
First, we construct a fixed probe from 32 test-set task inputs per intent, with paired store,
retrieve, revise, and remove intents. The \mbox{CAMG-RL} and AgeMem versions preserve the visible task prefix and memory content while
changing the interface description and executable action. A third version keeps the CAMG-RL filesystem
backend, argument schema, file paths and bytes, observations, operation semantics, action budget, and
executor unchanged, but renames \texttt{shell\_command} to the fixed nonce name
\texttt{tool\_kappa}. The parser maps this name back immediately before invoking
the same executor. Under frozen Qwen3.5-4B, we score the executable action tokens with the exact chat
template for each version. Action surprisal is their mean negative log probability in nats per token. We use the nonce-name variant only for the frozen-base
probe. A fourth probe variant, the hybrid variant, keeps the six AgeMem tools, their signatures, and their
semantics, but renders every call and response in the native function-call format, so the call
format is held constant while the memory-tool semantics vary. This variant is also scored only under the frozen base.
A fifth diagnostic variant deletes every shell-command documentation passage from the hybrid
variant's prompt, covering the schema entry, teaching lines, worked examples, and protocol paragraphs,
while holding the appended memory-tool block, target actions, setup turns, and responses
byte-identical to the hybrid variant. In the coding environment the deletion leaves the six
memory tools as the only documented functions. This variant is a counterfactual prompt, because in the
coding and AutoResearch environments the shell tool is also the task interface, and it is scored
only under the frozen base.

\paragraph{Token-level decomposition.}
A token-level decomposition of the same frozen-base items splits each target action into the tokens
of its tool name, located by byte offset in the action text through the tokenizer's offset map, and
all remaining tokens. Each paired gap to the CAMG-RL version then decomposes exactly into a name-token
contribution and a remainder contribution, each weighting its segment's nats by the target's own
token count, so the two contributions sum to the paired gap.

\paragraph{Training-quarter metrics.}
Second, we divide each 200-update training run into four equal quarters. A memory-operation attempt is a
policy response that names \texttt{.agent\_memory/} or one of AgeMem's six memory tools, including a
malformed call. The executable-operation rate is the fraction of these attempts that parse and execute. A
complete chain contains a successful store or revision, a later retrieval of the saved content, and a
later task action that uses a saved fact recovered from memory. The chain rate uses all episodes in
the quarter as its denominator. The chains therefore measure the memory use the policy adds beyond
the mandatory continuation write: every context replacement already requires a successful
\texttt{.agent\_memory/CONTINUATION.md} write (Appendix~\ref{app:state-action-contract}), and this
required write is not a chain link. In updates 181--200, 60.5\% of episodes cross at least one context
replacement and thus executes the required write. Among these episodes, 41.3\% also
store to a memory file beyond the continuation file, 71.6\% retrieve earlier memory content, and
34.8\% complete a full chain. From the per-update chain curve, $T_{50}$ and $T_{90}$ are the first
updates whose trailing ten-update mean reaches 50\% or 90\% of the mean over updates 181--200 and remains
above that level for the next ten updates.

\paragraph{Policy-shift distance.}
Third, on the fixed probe prefixes, we compute the token-averaged
$D_{\mathrm{KL}}(\pi_{\mathrm{trained}}\,\|\,\pi_{\mathrm{base}})$ between each trained policy and its
frozen initialization. This measures how far the action distribution moved to support each interface. In weight space,
the relative change $\|\Delta W\|/\|W\|$ over all language-model weights is 0.0022 for CAMG-RL
and 0.0019 for AgeMem, and the vision tower is byte-identical to the frozen base in both models.
The frozen-base probe isolates the action rendering.

\paragraph{AgeMem baseline under common training.}
For the AgeMem baseline, we retain the six LTM/STM tools from the original
method~\citep{yu2026agenticmemory} but use the same base checkpoint, CAMG task schedule, fully asynchronous
PPO, native task rewards, and 200-update budget as CAMG-RL. This design compares the core memory mechanisms
under common training. In the original AgeMem recipe, three-stage progressive training, step-wise GRPO,
and composite rewards are combined. Using that recipe here would change the curriculum, optimizer, and
reward together with the memory interface. Appendix~\ref{app:agemem-original-recipe} replicates the original recipe in full and leaves the ordering unchanged. The comparison still
bundles the storage backend with each interface's observation, retrieval, summarization, filtering, and
capacity semantics. Appendix~\ref{app:applypatch-interface} isolates the interface,
rerunning the same protocol with file modification routed through a single out-of-distribution
tool, and the resulting model trails the shell interface on every environment. The analysis attributes this hypothesis to the measured frozen-policy distribution without inferring which
training data produced it.

\paragraph{Interface-prior measurements.}
Under the frozen base, filesystem actions cost 1.193 nats per token on store intents, 0.076 on
retrieve, 0.252 on revise, and 0.272 on remove. The nonce-name control changes these surprisals by at most 0.013 nats per
token on any intent. AgeMem's dedicated tools cost more on every intent in their own call format, and
Table~\ref{tab:interface-gap-split} decomposes each paired gap into its name-token and remainder
contributions. For the dedicated tools the store gap sits entirely in the remainder ($-0.001$ nats
per token on the name tokens), so their cost lies in the action markup and payload, while on revise
and remove their positive net gaps come from the name tokens ($+0.038$ and $+0.165$). In the
hybrid variant the name tokens are the expensive positions, at 6.371 nats per token on store
against 0.252 for the native name, and they carry $+0.118$ of that variant's $+0.166$ store gap. Switching
the call format recovers only 0.039 of the 0.205 nats-per-token store gap and makes the retrieve gap
larger, so the frozen-base prior favors the filesystem action rendering itself, with both the tool
name and the call format measured as controls. The nonce-name control therefore bounds
the cost of renaming a documented tool, while an undocumented name in the native function-name slot
is itself expensive.

\paragraph{Shell-deletion measurements.}
Table~\ref{tab:interface-name-cost} reports the name-token surprisal of every
variant. Deleting the shell documentation from the hybrid variant's prompt leaves the foreign
names expensive: the store name tokens still cost 6.757 nats per token against 6.371 with the
documentation present, and 9.035 in the coding environment, where the deletion leaves the six
memory tools as the only documented functions. The paired store gap grows to $+0.354$ nats per token, of which the name tokens carry $+0.120$ and
the remainder $+0.234$, while the retrieve, revise, and remove gaps stay close to the hybrid
variant's gaps. The name-token cost therefore reflects the base policy's pre-trained prior over action
functions: a documented rename of the habitual filesystem function stays cheap, and an undocumented
name in the function-name slot stays expensive with the documented competitor either present or
deleted.

\paragraph{Training-dynamics and policy-shift measurements.}
During training,
CAMG-RL attempts memory operations in 18.7 to 30.2\% of responses across quarters, its
executable-operation rate rises from 72.8\% in the first quarter to 86.8\% in the last, and its
complete-chain rate rises from 7.97\% to 20.9\% of episodes with $T_{50} = 54$ and $T_{90} = 79$
updates. AgeMem attempts its memory tools 45, 6, 1, and 1 times in the four quarters (53 of
261{,}941 responses) and never completes a chain, so its $T_{50}$ and $T_{90}$ are not defined.
Of the 53 attempts, 49 fail to parse as valid calls, most often a mismatched call envelope, and
one well-formed call executes. The remaining three responses name a tool in prose without
emitting a call. CAMG-RL stores and reads memory through ordinary shell commands, so no
separate memory-call grammar exists to fail. The action parser rejects 1{,}863 of its
306{,}095 training responses (0.61\%), all on Shop (966, 448, 448, and 1 across the four
quarters), and 835 of the 76{,}446 memory attempts lie inside rejected responses and never
execute. The mean token-averaged $D_{\mathrm{KL}}$ from the trained policy to the
frozen base is 0.094 nats per token for CAMG-RL (0.216 on store, 0.075 on retrieve, 0.029 on revise,
and 0.058 on remove) and 0.064 for AgeMem (0.136, 0.046, 0.025, and 0.051). Both trained policies
remain close to the frozen base, and only the CAMG-RL policy acquires complete memory chains.
The prior measured on the frozen base reaches training through emission: filesystem actions are
produced and executed from the first quarter, while the six tools, more surprising on every intent
in their own call format, appear almost only in malformed form, and reinforcement can build only on
what the policy emits.

\subsection{External Evaluation}
\label{app:external}

We evaluate transfer on SWE-bench Verified \citep{jimenez2024swebench} and MLE-bench Lite
\citep{chan2024mlebench}. The terminal CAMG-RL checkpoint and the frozen models in
Table~\ref{tab:matched-external-methods} use the same tasks, decoding settings, policy-action budget,
environment runtime, and grader. Invalid outputs, exhausted budgets, timeouts, and missing submissions
count as failures. CAMG-RL retains its trained file-memory interface. At each context boundary, each frozen
comparator instead generates an automatic summary with a fixed prompt and byte limit. Summary calls do
not consume policy actions, but their calls, tokens, latency, and cost are reported. Neither benchmark is
used for training. The table reports one primary success rate per benchmark and their unweighted average.

\paragraph{SWE-bench Verified.}
\label{app:swebench}
The primary score is issue-resolution rate on all 500 tasks under the official container grader. A task
counts as successful only when the submitted patch passes that grader. Every method receives the same
repository snapshot, container image, shell and editing interface, action budget, and timeout.

\paragraph{MLE-bench Lite.}
\label{app:mlebench}
The primary score is Any Medal rate across all 22 competitions, namely the fraction of competitions in
which the terminal submission reaches at least the official bronze-medal threshold. Each method receives
the prepared public data and task description, then produces one terminal submission for the official
private-test grader. The grader's continuous competition score and the valid-submission rate are secondary
diagnostics and do not enter the cross-benchmark average.

\paragraph{Evaluation precision.}
Under exact binomial 95\% intervals, the SWE-bench Verified rates are CAMG-RL-4B $[12.7, 19.3]$,
CAMG-RL-9B $[23.7, 31.7]$, Qwen3.5-4B $[5.4, 10.3]$, the frozen 9B $[11.8, 18.2]$, the 27B $[14.2, 21.0]$,
the 35B-A3B $[12.5, 19.1]$, the 122B-A10B $[18.4, 25.9]$, and the 397B-A17B $[30.2, 38.7]$, so the
CAMG-RL-4B and 35B-A3B intervals overlap almost completely. On MLE-bench Lite one medal moves the Any-Medal rate by 4.5 points: the
one-medal models, CAMG-RL-4B among them, share $[0.1, 22.8]$, the two-medal models,
CAMG-RL-9B and the 122B-A10B, share $[1.1, 29.2]$, and the three-medal 397B-A17B has $[2.9, 34.9]$. Context replacement occurs 0.78
times per completed SWE-bench Verified trajectory, 385 replacement events across the 493 completed trajectories, and
0.59 times per MLE-bench Lite task, where 13 of 22 tasks cross exactly one boundary.
The remaining seven SWE-bench Verified tasks receive no resolution judgment, six because the
sandbox environment fails before the trajectory completes and one because the submitted patch is
too large for the official grader to classify. All seven count as failures in the 500-task
denominator, as do unjudged tasks for the comparison models. Across the CAMG test set,
65.0\% of episodes cross at least one context replacement, and replacement occurs 1.95 times per
episode (4.95 on Shop, 1.61 on DeepResearch, 0.88 on Coding, and 0.38 on AutoResearch).

\paragraph{Contamination control.}
No training task shares a repository or competition with the external benchmarks. The full SWE-smith
corpus behind the Coding environment spans 223 repositories and contains none of SWE-bench Verified's
12 repositories, so no Coding training task can come from an evaluated repository. Four training
repositories belong to organizations that also maintain SWE-bench Verified repositories (iniconfig,
astroid, click, and daphne), but each is a different codebase. All AutoResearch training families
are Kaggle dataset entities, and none of their identifiers matches any of MLE-bench Lite's 22
competitions.

%% file: sections/appendix_additional_results.tex
This section collects the token-level tables behind the interface-prior measurements in
Appendix~\ref{app:interface-prior-analysis}, together with the estimation control against the
default CAMG-RL on the same 512 test tasks. The closing subsections pair blank and shuffled
replacements of each trajectory's saved memory content against intact content, replicate
AgeMem under its original training recipe, rerun CAMG-RL with file modification routed
through a single apply\_patch tool, train the two strongest methods five times each, and report
the wall-clock time of the formal training runs.

\subsection{Interface-Prior Name-Token Analysis}
\label{app:interface-name-token-analysis}
We test whether the paired gap is an artifact of the tool names.
Table~\ref{tab:interface-gap-split} splits each paired gap into its name-token and remainder
contributions. On the store and retrieve intents, where the paired gaps are largest, the plain
variants owe almost nothing to their names, and the retrieve gap keeps most of its mass in the
remainder even in native form. Table~\ref{tab:interface-name-cost} measures the name tokens
directly. Deleting every shell-command documentation passage holds everything else
byte-identical yet leaves the foreign names expensive, so their cost does not come from
documenting the filesystem competitor. The prior therefore favors the filesystem action
itself, not its name.

\begin{table}[H]
\centering
\small
\renewcommand{\arraystretch}{1.08}
\caption{\textbf{The paired gap is not about the tool name.} Each paired gap is the
difference in mean NLL per action token against the CAMG-RL version on the same probe items. The
name-token and remainder contributions each weight their segment's nats by the target's own token
count, so the two contributions sum to the paired gap. The third variant of each group deletes every shell-command documentation passage from
the hybrid variant's prompt.}
\label{tab:interface-gap-split}
\begin{tabular}{@{}llccc@{}}
\toprule
\textbf{Intent} & \textbf{Probe variant} & \textbf{Paired gap} & \textbf{Name} & \textbf{Remainder} \\
\midrule
Store & AgeMem tools & $+0.205$ & $-0.001$ & $+0.207$ \\
 & AgeMem tools in native form & $+0.166$ & $+0.118$ & $+0.047$ \\
 & \quad shell documentation removed & $+0.354$ & $+0.120$ & $+0.234$ \\
\midrule
Retrieve & AgeMem tools & $+0.398$ & $+0.022$ & $+0.376$ \\
 & AgeMem tools in native form & $+0.484$ & $+0.193$ & $+0.291$ \\
 & \quad shell documentation removed & $+0.477$ & $+0.177$ & $+0.300$ \\
\midrule
Revise & AgeMem tools & $+0.020$ & $+0.038$ & $-0.018$ \\
 & AgeMem tools in native form & $-0.007$ & $+0.066$ & $-0.074$ \\
 & \quad shell documentation removed & $-0.011$ & $+0.069$ & $-0.080$ \\
\midrule
Remove & AgeMem tools & $+0.062$ & $+0.165$ & $-0.103$ \\
 & AgeMem tools in native form & $-0.017$ & $+0.234$ & $-0.251$ \\
 & \quad shell documentation removed & $-0.024$ & $+0.225$ & $-0.248$ \\
\bottomrule
\end{tabular}
\end{table}

\begin{table}[H]
\centering
\small
\renewcommand{\arraystretch}{1.08}
\caption{\textbf{Deleting the documented competitor leaves the foreign names expensive.} Each cell is the mean
surprisal in nats per token of the tool-name tokens of that variant's target action, located by byte
offset through the tokenizer's offset map. The first two rows call the documented filesystem tool
under its native and nonce names. The last row deletes every shell-command documentation passage
from the hybrid variant's prompt, holding the appended memory-tool block, target actions,
setup turns, and responses byte-identical, which leaves the six memory tools as the only
documented functions in the coding environment.}
\label{tab:interface-name-cost}
\begin{tabular}{@{}lcccc@{}}
\toprule
\textbf{Probe variant} & \textbf{Store} & \textbf{Retrieve} & \textbf{Revise} & \textbf{Remove} \\
\midrule
CAMG-RL files & 0.252 & 0.031 & 0.031 & 0.031 \\
Aliased files & 0.207 & 0.009 & 0.009 & 0.009 \\
AgeMem tools & 0.053 & 0.531 & 3.092 & 3.327 \\
AgeMem tools in native form & 6.371 & 3.763 & 5.184 & 5.306 \\
\quad shell documentation removed & 6.757 & 3.323 & 5.319 & 5.095 \\
\bottomrule
\end{tabular}
\end{table}

\subsection{Token-Level Estimation Control}
\label{app:token-level-estimation}
This control replaces the response-level advantage estimation of
Eq.~\eqref{eq:action-gae} with per-token GAE: every sampled token carries its own
temporal-difference recursion over the pre-token values $V^{-}_{t,j}$, and the resulting token
advantages pass through the same centering, environment token-count weighting, masked
standardization, and clipped PPO update as the default. Rewards, action and token budgets, prompts,
decoding, critic training, and the 200-update training schedule are unchanged, and the control is
evaluated on the same 512 test tasks with the same runtimes and graders.
Table~\ref{tab:token-level-estimation} reports the result next to the default CAMG-RL. The control
trails the default by 4.3 points on average. The deficit concentrates in Coding ($-7.8$),
DeepResearch ($-7.7$), and Shop ($-3.1$), while AutoResearch changes by $+1.6$ points in the
control's favor.

\begin{table}[H]
\centering
\small
\renewcommand{\arraystretch}{1.08}
\caption{\textbf{Token-level estimation control.} Success rates (\%) on the 512 test tasks of
Appendix~\ref{app:evaluation-details}. The control replaces response-level advantage estimation
with per-token GAE and changes nothing else in training or evaluation.}
\label{tab:token-level-estimation}
\begin{tabular}{@{}lccccc@{}}
\toprule
\textbf{Policy} & \textbf{Shop} & \textbf{Coding} & \textbf{DeepResearch} & \textbf{AutoResearch} &
\textbf{Average} \\
\midrule
CAMG-RL & 97.1 & 26.6 & 55.5 & 38.3 & 54.4 \\
\quad token-level estimation & 94.0 & 18.8 & 47.8 & 39.9 & 50.1 \\
\bottomrule
\end{tabular}
\end{table}

\subsection{Memory-Content Intervention}
\label{app:memory-content-intervention}
\begin{figure}[H]
  \centering
  \includegraphics[width=0.98\linewidth]{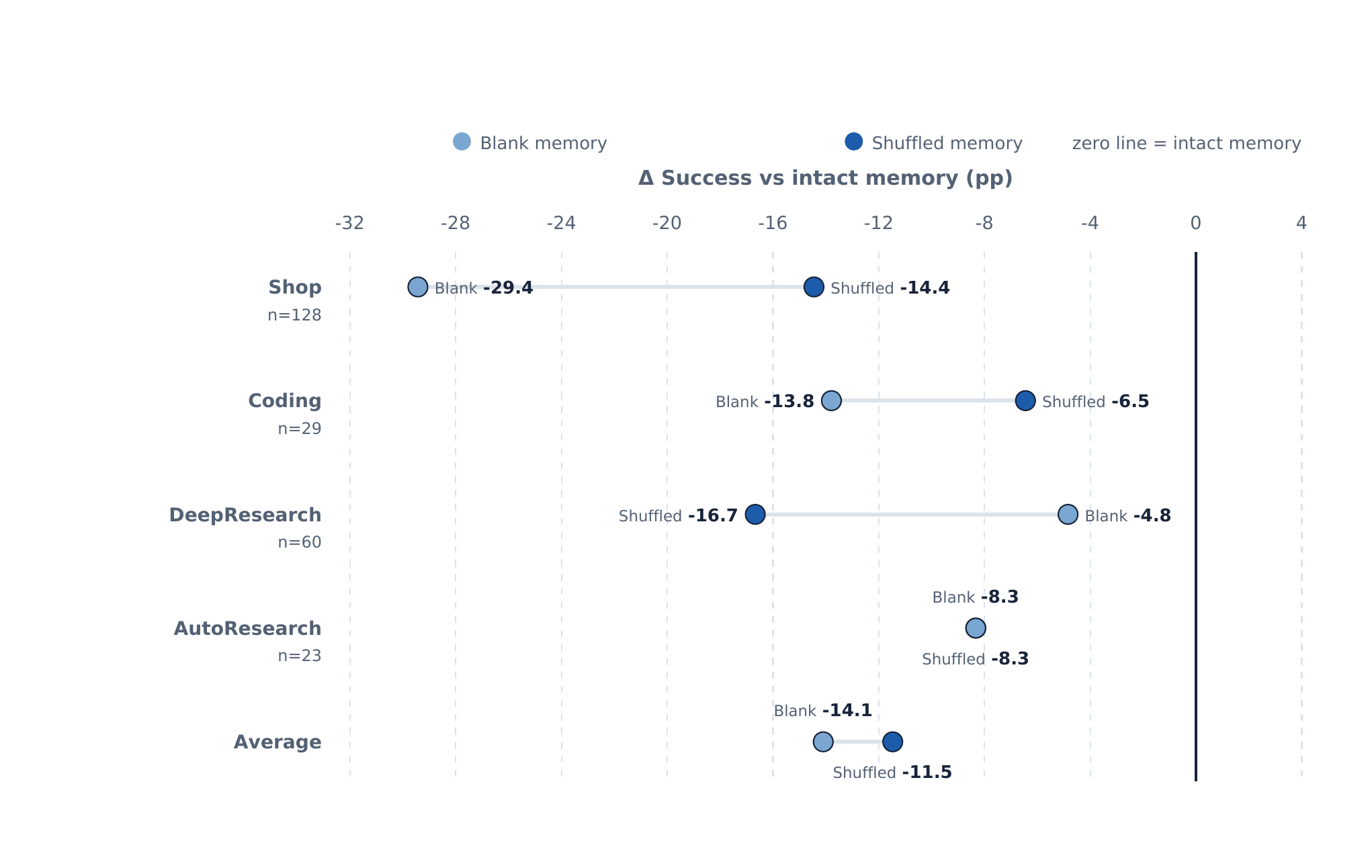}
  \caption{\textbf{Saved memory content drives later success.} Paired success-rate change from replacing
  each trajectory's memory files at its first context replacement: blank (erased) and shuffled
  (transplanted from a length-matched same-environment task) versus intact content, in percentage
  points across all four environments.}
  \label{fig:memory-intervention}
\end{figure}

We isolate the causal role of the saved content. Across all four environments, trained-policy
trajectories with nonempty memory are forked immediately after their first context replacement
into intact-memory, blank-memory, and same-environment shuffled-memory conditions
(Appendix~\ref{app:evaluation-protocols}). The task state, policy, sampling state, budget,
interface, and non-memory files remain fixed. The forking harness itself performs the required
post-reset re-read of the continuation checkpoint and withholds its observation, and all
non-policy steps and turn-budget skips are identical across conditions. Blanking the saved content
cuts average success by 14.1 points and transplanting task-mismatched content by 11.5
(Figure~\ref{fig:memory-intervention}). The largest blank drop is in Shop (-29.4), the
largest shuffled drop in DeepResearch (-16.7), and blank and shuffled content each reduce
success in every environment.

\subsection{AgeMem under Its Original Recipe}
\label{app:agemem-original-recipe}

The main-table AgeMem model trains under the common CAMG protocol to isolate the memory mechanism.
To test whether the original training recipe changes the comparison, we replicate that recipe in
full, with its three-stage progressive training, step-wise GRPO, and composite rewards, on the
same base checkpoint and CAMG environments, and evaluate it on the same test tasks
(Table~\ref{tab:agemem-recipe}). The recipe lifts Shop by 42.6 points and DeepResearch by 19.6 over the
common-training model, while Coding and AutoResearch land 1.6 and 7.0 points below it. The
original-recipe model completes the store--retrieve--use chain in 14.1\% of episodes, a behavior the
common-training model never shows, yet CAMG-RL still leads by 15.7 points on average and completes
chains in 20.9\% of episodes in its final training quarter. The ordering of
Table~\ref{tab:native-heldout} is therefore robust to the AgeMem training recipe.

\begin{table}[H]
\centering
\scriptsize
\setlength{\tabcolsep}{2.6pt}
\renewcommand{\arraystretch}{1.08}
\caption{\textbf{The original AgeMem recipe does not close the gap.} Success rate, \%, for the common-training AgeMem model of Table~\ref{tab:native-heldout}, the same six
tools trained with the full original recipe, and CAMG-RL. Chain is the share of episodes that complete
the store--retrieve--use chain in the final training quarter, and the common-training model never
completes one.}
\label{tab:agemem-recipe}
\begin{tabularx}{\linewidth}{@{}L{0.34\linewidth}*{6}{>{\centering\arraybackslash}X}@{}}
\toprule
& \multicolumn{5}{c}{\textbf{Success rate, \%}} & \\
\cmidrule(lr){2-6}
\textbf{Method} & \textbf{Shop} & \textbf{Coding} &
\shortstack{\textbf{Deep}\\\textbf{Research}} &
\shortstack{\textbf{Auto}\\\textbf{Research}} & \textbf{Average} & \textbf{Chain} \\
\midrule
AgeMem, common training & 16.7 & \textbf{26.6} & 28.1 & 29.7 & 25.3 & 0.0 \\
AgeMem, original recipe & 59.3 & 25.0 & 47.7 & 22.7 & 38.7 & 14.1 \\
\midrule
\textbf{CAMG-RL} & \textbf{97.1} & \textbf{26.6} & \textbf{55.5} & \textbf{38.3} & \textbf{54.4} & \textbf{20.9} \\
\bottomrule
\end{tabularx}
\end{table}

\subsection{The DeepResearch Drop of Common-Training AgeMem}
\label{app:agemem-deepresearch-drop}

Common-training AgeMem is the only learned-memory method in Table~\ref{tab:native-heldout} that
falls below its frozen base on an environment, accepting 28.1\% of DeepResearch answers against
the frozen base's 36.7\%. The drop is the out-of-distribution tool difficulty made concrete. In
the AgeMem interface, keeping an episode within the training-matched 30,720-token width falls to
the policy's own compaction action, and in 33 of the 128 test episodes that action never arrives
in time, so the episode is finalized at the environment horizon. The frozen base and CAMG-RL
keep width through the environment's context replacement, whose required policy cooperation is a
filesystem write inside the pre-training distribution, and they record no such episode. Why the
action never arrives is the training story. The common protocol pays only native task reward,
which reaches an interface behavior only through later task success, and the six tools start
outside the pre-training distribution, so emission is unreliable from the first update and never
consolidates. The chain measurement records zero completed store--retrieve--use chains in 200
updates, and the compaction action belongs to the same interface and consolidates no better.
Because success on
DeepResearch requires an accepted terminal answer, the 36 accepted answers all fall in the
remaining 95 episodes, a rate of 37.9\% in line with the frozen base, so the premature
finalizations, not weaker answering, account for the deficit. Thirty-three of the 34 finalizations
recorded across the four environments fall on DeepResearch, the frozen base's strongest
environment, while the other three start from frozen-base rates of 10.2\% to 11.6\% and gain
5.1, 15.7, and 19.5 points under the common protocol. The original recipe, whose
composite rewards train the tool behavior directly under three-stage progressive training,
recovers DeepResearch to 47.7\% (Table~\ref{tab:agemem-recipe}). The drop is the
training-side confirmation of the surprisal gap between the two interfaces. With the
native task reward, the environments, and the base policy fixed, moving the memory
actions outside the pre-training distribution is enough to pull the strongest
environment below its frozen base. Memory learnability under the native task reward
alone is a property of where the environment puts its memory actions, and the six
tools sit on the wrong side of that line.

\subsection{The Coding Failures of CAMG-RL}
\label{app:coding-failures}

CAMG-RL resolves 34 of the 128 Coding test tasks, and the composition of the 94 failures
locates the remaining headroom. Every failure is a task failure, with no infrastructure
exclusion. Fifty-eight of the 94, 61.7\%, submit a patch whose hidden tests fail, after a
median of 24.5 actions against 22 for the successes, with a single submission inside the
first ten. The working sequence completes and the fix is wrong, so this class sits in
code-fixing capability at 4B scale, not in the memory machinery. Thirty-five, 37.2\%,
exhaust the full 40-action budget without submitting. Training and evaluation share this cap,
and the continuation checkpoint is in place in 33 of the 35 when the budget ends, so the
longest episodes stop with their context boundaries crossed and their saved state intact.
Only one episode finishes without an executed action, on a response that hits the length
limit and fails to parse. Ninety-four of the 128 episodes cross a context replacement
boundary, and in every one of the 94 the policy writes its continuation checkpoint at
the boundary and reads it back after the replacement, in the 12 succeeding episodes
and the 82 failing ones alike. The behavior that end-to-end training builds on native
task reward executes at every boundary it meets on held-out tasks. Coding failures
therefore divide into wrong fixes and unfinished fixes, and the headroom on Coding is
fix quality and action budget, with the trained memory behavior already running in
full.

\subsection{A Single Out-of-Distribution Tool}
\label{app:applypatch-interface}

The AgeMem comparison changes the memory interface and the tool inventory at once. To isolate
the interface, we train a variant that replaces native shell file edits with a single Codex-style
apply\_patch tool, with the same base checkpoint and CAMG
environments, and evaluate it on the same test tasks (Table~\ref{tab:applypatch}). The tool
is powerful, and the variant must learn only one extra name and format. It still lands below
CAMG-RL on all four environments, 48.6\% versus 54.4\% on average, and after training 11.7\%
of its responses fail to parse, against 0.61\% across the shell interface's whole training run and
a single rejected response in its final quarter. Memory acquisition slows but does not
stop: the variant completes the store--retrieve--use chain in 15.6\% of episodes, below the shell
interface's 20.9\%, and AgeMem under common training never completes one. Out-of-distribution emission, not tool
capability, is the bottleneck: one such tool already carries parse friction that training does
not erase, and six cost AgeMem almost every attempt.

\begin{table}[H]
\centering
\scriptsize
\setlength{\tabcolsep}{2.6pt}
\renewcommand{\arraystretch}{1.08}
\caption{\textbf{A single out-of-distribution file tool does not recover the shell interface.}
Success rate, \%, for CAMG-RL and a variant trained under the same protocol with file modification
routed through a single Codex-style apply\_patch tool. Parse is the share of responses the
action parser rejects, over the whole training run for the shell interface and after training for the
apply\_patch variant. Chain is the share of episodes that complete the store--retrieve--use chain
in the final training quarter.}
\label{tab:applypatch}
\begin{tabularx}{\linewidth}{@{}L{0.34\linewidth}*{7}{>{\centering\arraybackslash}X}@{}}
\toprule
& \multicolumn{5}{c}{\textbf{Success rate, \%}} & & \\
\cmidrule(lr){2-6}
\textbf{Interface} & \textbf{Shop} & \textbf{Coding} &
\shortstack{\textbf{Deep}\\\textbf{Research}} &
\shortstack{\textbf{Auto}\\\textbf{Research}} & \textbf{Average} & \textbf{Parse} & \textbf{Chain} \\
\midrule
CAMG-RL, shell file edits & \textbf{97.1} & \textbf{26.6} & \textbf{55.5} & \textbf{38.3} & \textbf{54.4} & \textbf{0.61} & \textbf{20.9} \\
CAMG-RL, apply\_patch tool & 90.6 & 20.3 & 52.3 & 31.3 & 48.6 & 11.7 & 15.6 \\
\bottomrule
\end{tabularx}
\end{table}

\subsection{Robustness Across Training Runs}
\label{app:training-run-robustness}

Table~\ref{tab:native-heldout} reports one training run per learned method. We train each
method five times and evaluate every run across the CAMG test set. The four-environment
average is 55.3 with a standard deviation of 2.9 for CAMG-RL and 47.1
with 3.5 for CompactionRL, so the mean $\pm$ standard-deviation intervals, [52.4, 58.2]
against [43.6, 50.6], do not overlap, and the single-run averages of
Table~\ref{tab:native-heldout}, 54.4 and 48.0, fall inside them. The 8.2-point gap between
the means is more than twice the larger of the two standard deviations. The ordering of
Table~\ref{tab:native-heldout} is therefore robust to the training run.

\subsection{Training Time}
\label{app:training-time}
Table~\ref{tab:training-time} reports the wall-clock time of the formal training runs. All four models
train for 200 updates on the same six training GPUs, roughly 120 to 140
GPU-hours each, and every run finishes within about one day.

\begin{table}[H]
\centering
\small
\renewcommand{\arraystretch}{1.08}
\caption{\textbf{Every formal training run finishes within about one day.} All four use the
shared asynchronous-PPO configuration of Table~\ref{tab:joint-training-config} on the same six
training GPUs.}
\label{tab:training-time}
\begin{tabular}{@{}lcc@{}}
\toprule
\textbf{Method} & \textbf{Scale} & \textbf{Wall-clock (h)} \\
\midrule
CAMG-RL-4B & 4B & 19.6 \\
CompactionRL & 4B & 22.4 \\
AgeMem & 4B & 23.6 \\
CAMG-RL-9B & 9B & 23.2 \\
\bottomrule
\end{tabular}
\end{table}

%% file: sections/related_extended.tex
\section{Extended Related Work}
\label{app:extended-related-work}

\paragraph{Persistent external-memory systems.}
These systems preserve information for later retrieval.
\citet{lewis2020rag} condition generation on retrieved documents. Generative Agents
\citep{park2023generative} retain past observations for later reflection. MemGPT
\citep{packer2023memgpt} moves information between active and persistent storage through
recall and archival tiers. Mem0 \citep{chhikara2025mem0} maintains long-term user memories through inference-time updates.
\citet{zhou2026filesystemmemory} expose ordinary filesystem state as memory
through agent-managed directories. Librarian Agents \citep{librarianmodels2026} reorganize a corpus in
a filesystem before a frozen answering agent receives queries. PRO-LONG \citep{fox2026prolong}
preserves complete interaction histories as lossless program-searchable logs. CAMG-RL adds no
separate memory system: the task-acting coding agent itself maintains memory through ordinary
filesystem actions.\looseness=-1

\paragraph{Agent-memory benchmarks.}
These benchmarks test how agents retrieve, update, and use information across long interactions.
LongMemEval \citep{wu2025longmemeval} tests extraction, temporal reasoning, updating, and abstention on
chat histories. MemoryAgentBench \citep{hu2026memoryagentbench} measures incremental retrieval,
test-time learning, long-range understanding, and selective forgetting. MemoryArena
\citep{he2026benchmarking} evaluates whether agents preserve and apply information across
interdependent multi-session interactions. These benchmarks evaluate memory without training it.
CAMG instead makes cross-context memory trainable from executable task reward.

\paragraph{Inference-time context compression.}
These methods compress interaction history into bounded context at inference time.
PACE \citep{wei2026pace} presents selected historical chunks at multiple resolutions. HiAgent
\citep{hu2025hiagent} summarizes completed subgoals while keeping detailed records retrievable.
ReSum \citep{wu2025resum} periodically restarts search from the query and a summary, with optional
distillation and reinforcement-learning adaptation. These methods are fixed at inference time.
CAMG-RL instead learns what to keep from downstream task reward.

\paragraph{Learned context management.}
These methods make active-context edits explicit policy decisions.
Memory as Action \citep{zhang2026memact} learns to replace selected history through prune-and-write
operations from downstream reward. Sculptor \citep{li2026sculptor} decides when to summarize, hide,
restore, search, or fragment context. AgentFold \citep{ye2026agentfold} applies context folding under
supervised fine-tuning, while \citet{liu2026contexttool} provide a callable compressor for learned
context editing. ACM \citep{li2026acm} learns offloading and retrieval through dedicated tools,
\citet{yi2026adacom} train a manager to edit a frozen executor's context, and \citet{sun2026scaling}
learn branch-and-return actions. Their edits remain inside the active context. CAMG-RL instead
persists memory in files beyond it.

\paragraph{Memory-specific credit assignment.}
These methods assign learning signals to memory decisions.
Mem-T \citep{yue2026memt} combines tree-guided rewards with hindsight attribution, and Fine-Mem
\citep{ma2026finemem} applies evidence-anchored attribution to chunk rewards. Memory-R2
\citep{yan2026memoryr2} compares edits within a global objective through local rerollouts from shared
memory states. ECHO \citep{xie2026echo} credits reused evidence and selection actions with positive
outcomes. CAMG-RL uses no memory-specific credit assignment, learning file memory directly from
task reward.\looseness=-1

\enlargethispage{3\baselineskip}
\paragraph{Agentic reinforcement learning.}
These works develop the reasoning patterns, execution interfaces, and learning algorithms used by long-horizon agents.
ReAct \citep{yao2023react} interleaves reasoning traces with environment actions, while Language Agent
Tree Search \citep{zhou2024lats} unifies self-reflection and environment feedback at inference time.
SWE-agent \citep{yang2024sweagent} shows that software agents depend on navigation, editing, shell
use, and feedback formatting. Agent Lightning \citep{he2026agentlightning} converts
model-call histories into RL transitions. ArCHer \citep{zhou2024archer} combines utterance-level value
learning with token-level policy optimization, Group-in-Group Policy Optimization \citep{feng2025gigpo}
uses episode- and step-level relative advantages, WebRL \citep{qi2025webrl} trains agents through
online curricula, and RAGEN \citep{wang2025ragen} adds trajectory-level stabilization. Search-R1
\citep{jin2025searchr} trains search agents from outcome rewards, LiteResearcher
\citep{literesearcher2025} adds difficulty-aware local-web environments, and DeepResearcher
\citep{deepresearcher2025} operates on the live web. MLE-RL \citep{mlerl2025} drives iterative
machine-learning engineering from task reward, Frontis-MA1 \citep{yang2026frontisma1} learns
long-horizon execution-grounded operators, and AgentRL \citep{zhang2025agentrl} trains one
asynchronous policy across multiple task families. CAMG-RL applies one asynchronous policy across
four task families where memory across context boundaries is itself learned.\looseness=-1

\paragraph{Asynchronous reinforcement-learning systems.}
These systems trace the move from parallel actors to fully decoupled agentic-RL pipelines.
\citet{mnih2016asynchronous} train shared parameters with parallel actor--learners, and IMPALA
\citep{espeholt2018impala} corrects policy lag with off-policy updates in centralized learners. SEED
RL \citep{espeholt2020seedrl} runs batched inference on the learner and Sample Factory
\citep{petrenko2020samplefactory} optimizes single-machine sampling for throughput. HybridFlow
\citep{sheng2025hybridflow} supports RLHF with distributed execution primitives. Recent systems adapt
actor--learner separation to long language-model trajectories. AReaL \citep{fu2025areal} consumes
mixed-policy rollouts under controlled staleness, Relax \citep{zhang2026relax} uses policy staleness
as a system control, and RollArt \citep{gao2025rollart} distributes rollout stages with bounded weight
staleness. SAO \citep{hou2026sao} reduces off-policy drift with single-rollout sampling and
double-sided token-level importance clipping. \citet{gao2026unlocking} decouple long
search trajectories from model updates. \mbox{CAMG-RL} builds on
this line, running fully asynchronous PPO across the four environments' variable-duration
trajectories.\looseness=-1